%% file: main_arxiv.tex
\documentclass[a4paper,fleqn]{cas-dc}
\input{frontmatter/orcidguard}

\usepackage[authoryear,longnamesfirst]{natbib}
\usepackage[T1]{fontenc}
\usepackage[utf8]{inputenc}
\usepackage{booktabs}
\usepackage{multirow}
\usepackage{array}
\usepackage{amsmath,amssymb}
\usepackage{graphicx}
\usepackage{listings}   %
\usepackage{url}
\usepackage{pcatodo}

\makeatletter
\def\arxiv@allowed@a{P-59}
\def\arxiv@allowed@b{P-78}
\renewcommand{\TODO}[2]{%
  \def\arxiv@id{#1}%
  \ifx\arxiv@id\arxiv@allowed@a\else
  \ifx\arxiv@id\arxiv@allowed@b\else
    \PackageError{main_arxiv}{Unfilled placeholder #1 in the arXiv build}%
      {Fill it, or decide explicitly to allow it in main_arxiv.tex.}%
  \fi\fi}
\renewcommand{\TODOfloat}[2]{%
  \PackageError{main_arxiv}{Unbuildable float #1 in the arXiv build}{}}

\makeatother

\ExplSyntaxOn
\cs_set:Npn \__cas_foot: { \hfill \sffamily \small \thepage \hfill }
\cs_set:Npn \__first_foot: { \hfill \sffamily \small \thepage \hfill }
\ExplSyntaxOff
\makeatletter
\ps@cas
\makeatother

\input{tables/counts}

\newif\ifanonymized
\anonymizedfalse

\graphicspath{{figures/exported/}{figures/plots/}}

\begin{document}
\let\WriteBookmarks\relax
\def\floatpagepagefraction{1}
\def\textpagefraction{.001}

\shorttitle{Joint Gating and Hint Generation in One Compact Decoder}
\shortauthors{Talukder and Haque}

\title[mode=title]{When to Assist and What to Suggest: Joint Gating and
       Memory-Grounded Hint Generation in One Compact Decoder}

\input{authors}

\input{sections/00_abstract}

\input{frontmatter/keywords}

\maketitle

\input{sections_common}

\input{sections/11_implications}

\input{sections/12_threats}

\input{sections/10_conclusion}

\input{frontmatter/declarations}

\input{frontmatter/acknowledgements}

\section*{Funding}
\input{frontmatter/funding}

\printcredits

\bibliographystyle{cas-model2-names}
\bibliography{refs}

\clearpage
\onecolumn
\input{supplement/inline}
\suppinlinefloats
\renewcommand{\topfraction}{0.9}
\renewcommand{\bottomfraction}{0.9}
\renewcommand{\textfraction}{0.07}
\renewcommand{\floatpagefraction}{0.85}
\section*{Supplementary material}
\input{supplement/body}

\end{document}

%% file: frontmatter/orcidguard.tex
\ExplSyntaxOn
\RenewDocumentCommand \printorcid { }
{
  \seq_if_empty:NF \g_stm_orcid_seq
    {
      \group_begin:
        \tex_let:D \thefootnote \relax \footnotetext
        {
          \raggedright
          \textsc{orcid}(s):\c_space_token
          \seq_use:Nn \g_stm_orcid_seq { ;~ }
        }
      \group_end:
    }
}
\ExplSyntaxOff

%% file: tables/counts.tex
\newcommand{\nMarkers}{28,058}
\newcommand{\nTriggers}{3,356}
\newcommand{\nSilence}{24,702}
\newcommand{\nDialogues}{866}
\newcommand{\nTriggerRate}{11.96}
\newcommand{\nPerlDialogues}{292}
\newcommand{\nSodaDialogues}{268}
\newcommand{\nSyntheticDialogues}{306}
\newcommand{\nNoiseFloor}{0.0070}
\newcommand{\nGateGapEarly}{0.0124}
\newcommand{\nGateGapLate}{0.0129}
\newcommand{\nASLbeatsBestPHolm}{0.0060}
\newcommand{\nASLThresholdOptimismMean}{0.0033}
\newcommand{\nASLThresholdOptimismHigh}{0.0079}
\newcommand{\nPromptedSubsampleN}{2,000}
\newcommand{\nPromptedZeroShotF}{0.1381}
\newcommand{\nPromptedFewShotF}{0.1914}
\newcommand{\nPromptedFewShotAUROC}{0.5773}
\newcommand{\nEqualisationContextDelta}{+0.0020}
\newcommand{\nEqualisationContextPHolm}{0.0472}
\newcommand{\nEqualisationRecipeDelta}{+0.0650}

\newcommand{\nEqualisationArchDelta}{+0.0519}

\newcommand{\nLossAblationWorstPHolm}{0.117}
\newcommand{\nFocalGammaTwoDelta}{\ensuremath{-}0.0310}

\newcommand{\nFocalAlphaFourDelta}{\ensuremath{-}0.0278}

\newcommand{\nTransferAurocFineTuned}{0.8349}
\newcommand{\nTransferAurocBaseline}{0.8291}

\newcommand{\nTransferAurocRefitDelta}{0.0001}
\newcommand{\nTransferExcludeFFineTuned}{0.2024}
\newcommand{\nTransferExcludeFBaseline}{0.2063}
\newcommand{\nProtocolBiasCorrectionKSix}{\ensuremath{-}0.0012}
\newcommand{\nProtocolBiasCorrectionKTen}{+0.0001}
\newcommand{\nSpeedupKSix}{1.76}
\newcommand{\nSpeedupKTen}{1.31}
\newcommand{\nTransferTriggerRate}{2.98}
\newcommand{\nTransferKappa}{0.361}
\newcommand{\nTransferKappaCeiling}{0.694}
\newcommand{\nSweepPoints}{182}

\newcommand{\nRagConfigs}{45}

\newcommand{\nJudgedItems}{200}
\newcommand{\nRubricItems}{199}
\newcommand{\nJudgeRubricCorr}{0.127}
\newcommand{\nContainVerbatim}{50.95}

\newcommand{\nContainment}{51.19}
\newcommand{\nFactsPerProfile}{11.99}
\newcommand{\nRagTriggers}{3,356}
\newcommand{\nResidentWeights}{1.05}
\newcommand{\nSecondAdapterMB}{42}
\newcommand{\nSecondBaseMB}{957}
\newcommand{\nSharedTrunkSavingMB}{915}
\newcommand{\nInvocationCut}{7.62}
\newcommand{\nSpeedupEightB}{4.39}
\newcommand{\nSpeedupOneB}{1.38}
\newcommand{\nGateShare}{82}
\newcommand{\nEndToEndPrecision}{0.2387}
\newcommand{\nEndToEndExactMatch}{0.2619}
\newcommand{\nEmitted}{3,682}

\newcommand{\nSpuriousShare}{23.5}
\newcommand{\nEndToEndPrecisionOnePct}{0.3011}
\newcommand{\nEndToEndExactMatchOnePct}{0.2122}
\newcommand{\nMemoryEffect}{24.7}
\newcommand{\nMemoryEffectPrecise}{24.73}
\newcommand{\nMemBertWith}{0.57}
\newcommand{\nMemBertWithout}{0.35}
\newcommand{\nMemRougeWith}{0.48}
\newcommand{\nMemRougeWithout}{0.13}
\newcommand{\nMismatchLengthSharePct}{\ensuremath{-}1.33}
\newcommand{\nMismatchVsNoneP}{0.343}

\newcommand{\nMemoryEffectRetrainedBaseline}{23.63}

\newcommand{\nScaleTieP}{0.79}
\newcommand{\nScaleTieEndToEndP}{0.338}
\newcommand{\nQwenCeiling}{35.79}
\newcommand{\nThresholdCollapse}{28.0}
\newcommand{\nThresholdCollapseModel}{Gemma-3-270M}

\newcommand{\nBaselineSpread}{0.012}
\newcommand{\nBaselineShift}{0.70}
\newcommand{\nRescaleVsSpread}{57}
\newcommand{\nPromptWithMem}{676}
\newcommand{\nPromptNoMem}{370}
\newcommand{\nPromptMemCost}{306}
\newcommand{\nDialogueCap}{1,300}
\newcommand{\nCpuGateNfFourKSix}{1,793}

\newcommand{\nCpuHintNfFourKSix}{7.79}
\newcommand{\nCpuAgreeNfFourKSix}{0.998}
\newcommand{\nCpuDecisionNfFourKSix}{100.0}
\newcommand{\nCpuHintMatchNfFourKSix}{90}
\newcommand{\nCpuSpeedupNfFourKSix}{2.86}
\newcommand{\nCpuSpeedupOrtFullKSix}{2.72}
\newcommand{\nCpuSpeedupPtFullKSix}{1.75}
\newcommand{\nCpuIdealSpeedupKSix}{2.67}
\newcommand{\nCpuGateNfFourKTen}{2,909}

\newcommand{\nCpuHintNfFourKTen}{5.01}
\newcommand{\nCpuAgreeNfFourKTen}{0.991}
\newcommand{\nCpuDecisionNfFourKTen}{96.7}
\newcommand{\nCpuHintMatchNfFourKTen}{95}
\newcommand{\nCpuSpeedupNfFourKTen}{1.51}
\newcommand{\nCpuSpeedupOrtFullKTen}{1.54}
\newcommand{\nCpuSpeedupPtFullKTen}{1.35}
\newcommand{\nCpuIdealSpeedupKTen}{1.60}
\newcommand{\nCpuGateOrtFullKSix}{1,611}
\newcommand{\nCpuGatePtFullKSix}{4,330}
\newcommand{\nCpuGatePtBfKSix}{22,054}
\newcommand{\nCpuGatePtHalfKSix}{17,683}
\newcommand{\nCpuHintMarkers}{20}
\newcommand{\nCpuHostKinds}{2}
\newcommand{\nCpuDiskFull}{4.99}
\newcommand{\nCpuDiskNfFour}{1.65}
\newcommand{\nCpuHostsRecorded}{6}

\newcommand{\nHumanRatings}{1,144}
\newcommand{\nHumanSilenceAnswers}{843}
\newcommand{\nHumanDialogues}{32}
\newcommand{\nHumanDialoguesPlanned}{100}
\newcommand{\nHumanWhispersPlanned}{433}
\newcommand{\nHumanAnnotatorsPlanned}{5}

\newcommand{\nHumanAnnOne}{829}
\newcommand{\nHumanAnnOneOf}{833}
\newcommand{\nHumanAnnFive}{313}
\newcommand{\nHumanAnnThree}{2}
\newcommand{\nHumanAnnFiveConvs}{10}
\newcommand{\nJudgeFreeMemAgnes}{3.09}
\newcommand{\nJudgeFreeNoMemAgnes}{3.29}
\newcommand{\nJudgeFreeMemDeep}{3.16}
\newcommand{\nJudgeFreeNoMemDeep}{3.70}
\newcommand{\nHumanAlpha}{0.72}
\newcommand{\nHumanAlphaUnits}{196}
\newcommand{\nHumanMargin}{0.25}
\newcommand{\nHumanHoneMem}{4.01}
\newcommand{\nHumanHoneNoMem}{3.18}
\newcommand{\nHumanHoneDiff}{+0.83}
\newcommand{\nHumanHoneLo}{+0.48}
\newcommand{\nHumanHoneHi}{+1.23}
\newcommand{\nHumanUSixOneBDiff}{+0.04}
\newcommand{\nHumanUSixOneBLo}{$-$0.31}
\newcommand{\nHumanUSixOneBHi}{+0.41}
\newcommand{\nHumanUTenOneBDiff}{$-$0.02}
\newcommand{\nHumanUTenOneBLo}{$-$0.35}
\newcommand{\nHumanUTenOneBHi}{+0.30}
\newcommand{\nHumanUSixEightBDiff}{$-$0.38}
\newcommand{\nHumanUSixEightBLo}{$-$0.69}
\newcommand{\nHumanUSixEightBHi}{$-$0.04}
\newcommand{\nHumanUTenEightBDiff}{$-$0.44}
\newcommand{\nHumanUTenEightBLo}{$-$0.83}
\newcommand{\nHumanUTenEightBHi}{$-$0.07}
\newcommand{\nHumanEightOverOneB}{+0.42}
\newcommand{\nHumanOneBMean}{3.60}
\newcommand{\nHumanRetestOneB}{92.1}
\newcommand{\nHumanRetestEightB}{91.8}

\newcommand{\nHumanRenderings}{7}
\newcommand{\nHumanReconBelowEm}{4}
\newcommand{\nHumanRenderingsCorrelated}{5}
\newcommand{\nHumanCorrNativeEight}{0.55}
\newcommand{\nHumanCorrEmEight}{0.31}
\newcommand{\nHumanCorrRougeEight}{0.51}
\newcommand{\nHumanCorrBertEight}{0.47}
\newcommand{\nHumanCorrNativeOne}{0.67}
\newcommand{\nHumanCorrEmOne}{0.43}
\newcommand{\nHumanCorrRougeOne}{0.62}
\newcommand{\nHumanCorrBertOne}{0.61}
\newcommand{\nMemGateF}{0.7981}
\newcommand{\nMemGateBaseF}{0.7926}
\newcommand{\nMemGateGain}{+0.0055}
\newcommand{\nMemGateGainLo}{$-$0.0010}
\newcommand{\nMemGateGainHi}{+0.0125}
\newcommand{\nMemGateUTenLead}{+0.0183}
\newcommand{\nMemGateUTenLeadLo}{+0.0106}
\newcommand{\nMemGateUTenLeadHi}{+0.0259}
\newcommand{\nMemGateUSixLead}{+0.0041}
\newcommand{\nMemGateUSixLeadLo}{$-$0.0040}
\newcommand{\nMemGateUSixLeadHi}{+0.0121}
\newcommand{\nMemGateUSixAurocDiff}{$-$0.0014}
\newcommand{\nMemGateUTenAurocDiff}{+0.0019}
\newcommand{\nMemGateShare}{23}
\newcommand{\nMemGateShareLo}{$-$5}
\newcommand{\nMemGateShareHi}{+49}
\newcommand{\nTwoStageGateF}{0.8002}
\newcommand{\nTwoStageThreshold}{0.60}
\newcommand{\nTwoStagePrecision}{0.7648}
\newcommand{\nTwoStageRecall}{0.8391}

\newcommand{\nValMarkers}{26,410}
\newcommand{\nValPrevalence}{12.81}

\newcommand{\nFinalStep}{53,772}
\newcommand{\nUnifiedEpochs}{2}
\newcommand{\nTwoStageEpochs}{4}
\newcommand{\nSeedStep}{21,703}

\newcommand{\nUnifiedGateFKSix}{0.8022}
\newcommand{\nUnifiedGateGapKSix}{+0.0020}
\newcommand{\nSeedMeanKSix}{0.7846}
\newcommand{\nSeedSdKSix}{0.0033}
\newcommand{\nSeedGapKSix}{$-$0.0156}
\newcommand{\nUnifiedEndToEndEmKSix}{0.2929}

\newcommand{\nUnifiedGateFKTen}{0.8164}
\newcommand{\nUnifiedGateGapKTen}{+0.0162}
\newcommand{\nSeedMeanKTen}{0.8022}
\newcommand{\nSeedSdKTen}{0.0029}
\newcommand{\nSeedGapKTen}{+0.0020}
\newcommand{\nUnifiedEndToEndEmKTen}{0.2908}

\newcommand{\nEndToEndExactMatchEightB}{0.2548}

\newcommand{\nOracleCeilingOneB}{0.2971}
\newcommand{\nOracleCeilingEightB}{0.2947}

\newcommand{\nEndToEndExactMatchQwen}{0.3084}
\newcommand{\nOracleCeilingQwen}{0.3579}
\newcommand{\nSpeedupEightBKSix}{3.41}
\newcommand{\nTwoStageTestSelectedF}{0.8008}
\newcommand{\nSelectionMeanDelta}{0.0016}
\newcommand{\nSelectionWorstDelta}{0.0041}
\newcommand{\nSelectionSystems}{21}

%% file: authors.tex
\author[1]{A. J. A. Jubayer Talukder}[orcid=0009-0000-8450-254X]
\cormark[1]
\ead{ajajubayertalukder@gmail.com}
\credit{Conceptualization, Methodology, Software, Validation, Formal analysis,
Investigation, Data curation, Writing -- original draft, Visualization}

\author[1]{Mohammad Ariful Haque}[orcid=0000-0001-5555-4180]
\ead{arifulhoque@eee.buet.ac.bd}
\credit{Conceptualization, Resources, Writing -- review \& editing, Supervision,
Project administration}

\affiliation[1]{organization={Department of Electrical and Electronic Engineering
            (EEE), Bangladesh University of Engineering and Technology (BUET)},
            city={Dhaka},
            postcode={1205},
            country={Bangladesh}}

\cortext[1]{Corresponding author}

%% file: sections/00_abstract.tex
\begin{abstract}
Proactive assistants offer short, memory-grounded hints during a conversation, deciding
when to assist and what to suggest. Current systems split this between a gate and a
larger generator; we evaluate both together on \nMarkers{} decision points. Stage-wise
generation scores are oracle-gate ceilings: behind a real gate, end-to-end exact match is
\nEndToEndExactMatch{} against a ceiling of \nOracleCeilingOneB{}, and end-to-end
precision is \nEndToEndPrecision{}, leaving substantial headroom. A gate pays off through
precision and compute, whose saving falls from \nSpeedupEightB{}$\times$ with an 8B
generator to \nSpeedupOneB{}$\times$ with a 1B one. The memory profile drives hint
quality: removing it costs \nMemoryEffect{} exact-match points (rescaled BERTScore
\nMemBertWith{} to \nMemBertWithout{}), while retrieval and a larger generator add nothing
on exact match. One 1B decoder can do both, gating from an intermediate layer and
generating in the same pass. At its final checkpoint it matches or exceeds the two-model
pipelines on automatic end-to-end measures with \nSpeedupKTen{}--\nSpeedupKSix{}$\times$
less wall-clock time than the matched 1B pipeline; across three seeds, its deeper exit
ties the separate gate. A human evaluation (\nHumanRatings{} ratings over
\nHumanDialogues{} dialogues) supports the memory effect and rates the single model level
with the matched 1B pipeline and below the 8B one. Graded metrics and LlamaPIE's native
judge track human ratings more closely than exact match, which we report as a strict
lower bound.
\end{abstract}

%% file: frontmatter/keywords.tex
\begin{keywords}
proactive conversational assistants \sep intervention gating \sep memory-grounded
generation \sep early-exit decoding \sep large language model evaluation
\end{keywords}

%% file: sections_common.tex
\input{sections/01_introduction}

\input{sections/02_related}

\input{sections/03_system}

\input{sections/03_task_protocol}

\input{sections/04_gating}

\input{sections/05_generation}

\input{sections/06_unified}

\input{sections/07_endtoend}

\input{sections/08_evaluation}

%% file: sections/01_introduction.tex
\section{Introduction}
\label{sec:intro}

A proactive conversational assistant listens to a conversation the user is already
having and occasionally offers a short, memory-grounded hint: a name the user is
reaching for, a date, a topic worth raising. It is never invoked, so it must decide both
when to assist and what to suggest. Most candidate moments do not need a hint; about
\nTriggerRate{}\% do on our corpus. The two decisions depend on each other: a hint is
useful only at a moment that needs one, and a moment warrants an interruption only if a
useful hint exists.

\begin{figure*}[width=\textwidth,pos=t]
\centering
{\raggedright\small\textbf{(a)} What the wearer experiences\par}\smallskip
\includegraphics[width=\textwidth]{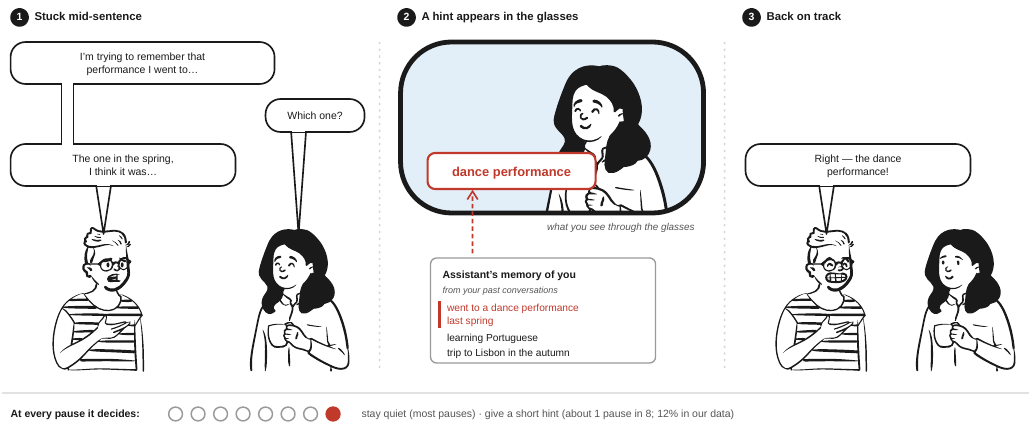}\par\medskip
{\raggedright\small\textbf{(b)} How the system does it\par}\smallskip
\includegraphics[width=\textwidth]{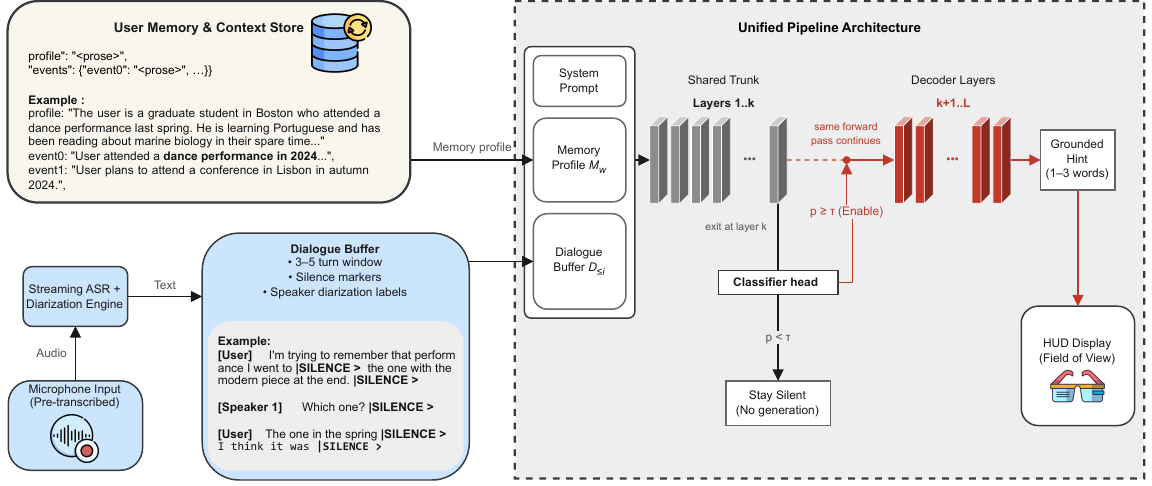}
\caption{The task and the system, on one illustrative example (not a corpus item).
(a)~At every pause the assistant decides whether to speak; it stays quiet at most pauses
(\nTriggerRate{}\% of pauses need a hint on our corpus) and otherwise shows a
one-to-three-word hint drawn from what it knows about the wearer. (b)~The dialogue
buffer and the memory profile form one prompt; a classifier head at layer $k$ decides
whether to speak (threshold $t$), and if so the same forward pass continues through layers
$k{+}1..L$ to the hint. Only the dashed box is trained and evaluated. People in (a)
adapted from Open Peeps (Pablo Stanley, CC0).}
\label{fig:overview}
\end{figure*}

Existing systems split the two decisions across two models, following
LlamaPIE~\citep{chen2025llamapie}: a small gate, and a larger generator invoked only when the gate fires.
We evaluate the two stages together over \nMarkers{} decision points. Stage-wise
generation scores turn out to be oracle-gate ceilings: behind a real gate, end-to-end
exact match is \nEndToEndExactMatch{} against a ceiling of \nOracleCeilingOneB{}, and
end-to-end precision is \nEndToEndPrecision{}, leaving substantial headroom. A gate
therefore pays off through precision and compute, and its compute saving depends on how
large the generator is.

Hint quality comes from the persistent user memory profile. Removing it costs
\nMemoryEffect{} exact-match points (rescaled BERTScore \nMemBertWith{} with the profile,
\nMemBertWithout{} without), while a larger generator of the same family and retrieval
over the profile add nothing on exact match.

These results motivate one model instead of two. We build a single 1B decoder that reads
its gate decision from an intermediate layer and, when the gate fires, continues the same
forward pass to generate the hint (Figure~\ref{fig:overview}b). Unlike confidence-based
early exit, the intermediate readout does not approximate the deep output; it decides
whether the deep computation runs at all. At its final checkpoint the unified model
matches or exceeds the two-model pipelines on the automatic end-to-end measures, with
\nSpeedupKTen{}--\nSpeedupKSix{}$\times$ less wall-clock time than the matched 1B
pipeline; across three seeds, its deeper exit ($k=10$) ties the separate gate.

A human evaluation (\nHumanRatings{} ratings over \nHumanDialogues{} dialogues) supports
the memory effect and rates the unified model level with the matched 1B pipeline and
below the 8B one. Graded metrics and LlamaPIE's native judge track the human ratings more
closely than exact match. We keep exact match as the primary automatic metric because it
alone has an end-to-end form, and read it as a strict lower bound on usefulness.

\paragraph{Contributions.}
\begin{itemize}
\itemsep0em
\item An end-to-end formulation that scores the composed gate-and-generate decision by
the triggers it serves and the precision of what it emits (Equation~\eqref{eq:e2e};
Sections~\ref{sec:problem} and~\ref{sec:endtoend}).
\item Evidence that the memory profile, not generator scale or retrieval, drives hint
quality, with controls that attribute the effect to its content
(Section~\ref{sec:generation}).
\item A single decoder that gates from an intermediate layer and generates in the same
pass, matching the two-model pipeline on automatic measures at lower cost
(Sections~\ref{sec:architecture}, \ref{sec:unified} and~\ref{sec:endtoend}).
\item A human evaluation that identifies which automatic signals track people
(Section~\ref{sec:evaluation}).
\end{itemize}

%% file: sections/02_related.tex
\section{Related work and positioning}
\label{sec:related}

\subsection{Proactive and mixed-initiative dialogue}

Proactive dialogue asks a system to contribute without being addressed, which makes
whether to speak part of dialogue management~\citep{deng2023proactive}. Recent
evaluation work treats the choice to intervene as a scorable act in its own
right~\citep{liu2025proactiveeval}. We follow that line and score the intervention
decision and the utterance \emph{jointly}, because a hint is useful only at a moment that
needs one.

Several recent assistive systems make the same commitment in wearable settings.
PROASSIST generates assistant dialogue from streaming egocentric
video~\citep{zhang2025proassist}; Pro$^2$Assist tracks procedural progress to time its
interventions~\citep{xu2026proassist}; Alpha-Service organises perception, memory and
interaction into a service architecture on AI glasses~\citep{wen2025alphaservice};
SocialMind delivers in-situ social suggestions through AR~\citep{yang2025socialmind};
Sensible Agent chooses how intrusively to offer help~\citep{sensibleagent2025}; and
Satori models user belief and intention to decide when guidance is
due~\citep{li2024satori}. Always-on but user-initiated systems such as
VisionClaw~\citep{liu2026visionclaw} have no intervention decision and fall outside this
comparison. Table~\ref{tab:positioning} positions these systems: the field is mostly
visual and wearable, ours is the text-only entry, and on-device evaluation is rarely
reported, including by us (our CPU timings, Section~\ref{sec:cpucost}, come from a cloud
container).

\input{tables/tab12_positioning}

\subsection{The two-stage baseline}

Our baseline reimplements the two-stage architecture of
LlamaPIE~\citep{chen2025llamapie} under our protocol: a $\leq$1B gating decoder followed
by a separately fine-tuned generator conditioned on the dialogue and the user memory
profile. Corpus preparation, adapter configuration and quantisation differ from the
original, so this is a controlled reimplementation, not a reproduction. LlamaPIE's
reported gate F1 and rubric scores serve as external reference points only: their
per-marker predictions are unavailable and their evaluation applies $\pm$1-turn
leniency, so no paired test against them is possible.

\subsection{Turn-taking and the modality boundary}

A mature literature predicts \emph{when} a speaker will yield the floor, from voice
activity projection~\citep{ekstedt2022vap} to full-duplex systems that manage
overlap~\citep{wang2026fullduplex}, and device-directed speech detection asks whether an
utterance was addressed to the assistant~\citep{wagner2023device}. These are natural
comparison points for a gate, but they consume audio and this corpus is text, so they
cannot be run here.

\subsection{Early exit, cascades and shared trunks}

Early-exit classifiers attach heads to intermediate layers and stop when a prediction is
confident enough~\citep{xin2020deebert}, and the idea extends to generation by exiting
per token~\citep{schuster2022calm}. Cascades save computation across models instead,
routing each query to the cheapest model likely to handle it~\citep{chen2023frugalgpt};
the two-stage assistant pipeline is such a cascade. Our shared adapter follows standard
low-rank adaptation~\citep{hu2022lora}.

In confidence-based early exit, the intermediate classifier approximates the full-depth
output. Our intermediate readout makes a different decision: whether the deep output
should be computed at all. The router and the expert are therefore the same weights, and
the routing decision is a prefix of the computation it authorises. Early exit is the
mechanism that makes this possible; it is not itself our claim.

\subsection{LLM-as-judge}

Reference-free evaluation by a strong model is standard for open-ended generation, with
reported agreement with human preference high enough to justify its use as a
proxy~\citep{zheng2023judging}. Section~\ref{sec:evaluation} tests that proxy against
human ratings on this task and identifies which judge to use.

%% file: tables/tab12_positioning.tex
\begin{table*}[t]
\centering
\footnotesize
\setlength{\tabcolsep}{2.0pt}
\begin{tabular}{>{\raggedright\arraybackslash}p{62pt}>{\raggedright\arraybackslash}p{54pt}>{\raggedright\arraybackslash}p{56pt}>{\raggedright\arraybackslash}p{58pt}>{\raggedright\arraybackslash}p{62pt}>{\raggedright\arraybackslash}p{60pt}>{\raggedright\arraybackslash}p{44pt}>{\raggedright\arraybackslash}p{54pt}}
\toprule
System & Modality & Model scale & Dataset & Gating mechanism & Generation eval. & On-device eval. & User study \\
\midrule
LlamaPIE \citep{chen2025llamapie} & speech (in-ear) & 1B gate + 8B gen. & semi-synthetic dialogue & separate small classifier & LLM rubric & Apple M2 & yes \\
PROASSIST \citep{zhang2025proassist} & streaming egocentric video & Llama-3.1-8B + SigLIP & synthesised from annotated video & end-to-end, learned & automatic, human-validated & \textit{not reported} & annotators only, no end users \\
Pro$^2$Assist \citep{xu2026proassist} & multimodal egocentric (AR) & Qwen3-VL-4B (2--30B swept) & 1,089 samples, 10 tasks & step-aware continuous reasoning & action understanding acc. & AR glasses & yes, n=20 \\
Alpha-Service \citep{wen2025alphaservice} & egocentric video (AI glasses) & Qwen2.5-VL-3B/7B + Qwen3-8B & case studies, no benchmark & VLM judges when to help & \textit{not reported} & AI glasses & \textit{not reported} \\
SocialMind \citep{yang2025socialmind} & multimodal AR sensors & GPT-4o / Llama-3.1 & 3 public corpora, 50 each & proactive update mechanism & LLM-as-judge, 3 custom dims. & AR glasses & yes, n=20 \\
Sensible Agent \citep{sensibleagent2025} & egocentric camera + multimodal & GPT-4o & 937 annotations, 40 participants & context-driven what/how selection & interaction effort, usability & XR headset & yes, n=10 \\
Satori \citep{li2024satori} & multimodal AR & GPT-4V / GPT-4 ensemble & WikiHow task descriptions & belief-desire-intention model & user ratings, NASA-TLX & no; offloaded to a server & yes, n=16 \\
\midrule
This work & text transcript & 1B, single shared adapter & semi-synthetic, \nMarkers{} markers & early exit at layer $k$, shared weights & end-to-end EM + precision, LLM judge, human uptake rating & \textit{not reported} & dialogue-level uptake rating (\nHumanRatings{} ratings) \\
\bottomrule
\end{tabular}
\caption{Proactive assistive systems compared. \textit{not reported}: the cited work does not report it.}
\label{tab:positioning}
\end{table*}

%% file: sections/03_system.tex
\section{Problem formulation and system architecture}
\label{sec:system}

\subsection{Problem definition}
\label{sec:problem}

A proactive assistant observes a conversation it was not invoked into. Let
$D = \langle u_1, \ldots, u_T \rangle$ be the dialogue observed so far and $M_w$ a
persistent memory profile for wearer $w$, accumulated before the conversation began. At
each candidate intervention point $i$ --- a silence between utterances --- the system
decides two things:

\begin{enumerate}
\itemsep0em
\item a \emph{gate} value $g_i \in \{0, 1\}$, whether to intervene, given the left
context $D_{\leq i}$. A gate model emits a score $p_i \in [0,1]$ and fires at a
threshold $t$:
\begin{equation}
g_i(t) = \mathbf{1}\left[\, p_i \geq t \,\right];
\label{eq:gate}
\end{equation}
\item when $g_i = 1$, a \emph{hint} $h_i$ of one to three tokens, generated from
$D_{\leq i}$ and $M_w$.
\end{enumerate}

A false alarm and a miss cost the wearer different things. A false alarm interrupts a
conversation that did not need it; a miss leaves the wearer as they would be without the
system. Because this asymmetry is a property of the deployment, we report gate quality
against a false-alarm budget instead of at one fixed operating point, and treat the
budget as a deployment decision (Section~\ref{sec:implications}).

Two end-to-end quantities score the composed decision. Let $\mathcal{T}$ be the set of
reference triggers --- the points that needed a hint --- and
$m_i = \mathbf{1}[h_i = h_i^{*}]$ indicate that the hint generated at point $i$ exactly
matches the reference $h_i^{*}$. Then
\begin{equation}
\mathrm{EM}_{\mathrm{e2e}} = \frac{1}{|\mathcal{T}|} \sum_{i \in \mathcal{T}} g_i\, m_i,
\qquad
\mathrm{P}_{\mathrm{e2e}} = \frac{\sum_{i \in \mathcal{T}} g_i\, m_i}{\sum_{i} g_i}.
\label{eq:e2e}
\end{equation}
\emph{End-to-end exact match} is recall-type: a suppressed trigger scores zero and a false
alarm does not enter it. \emph{End-to-end precision} divides by every emitted hint, false
alarms included. The first factors as
\begin{equation}
\begin{gathered}
\mathrm{EM}_{\mathrm{e2e}} = R \cdot \mathrm{EM}_{\mathrm{fwd}}, \\
R = \frac{1}{|\mathcal{T}|}\sum_{i \in \mathcal{T}} g_i,
\qquad
\mathrm{EM}_{\mathrm{fwd}} = \frac{\sum_{i \in \mathcal{T}} g_i\, m_i}{\sum_{i \in \mathcal{T}} g_i},
\end{gathered}
\label{eq:decomp}
\end{equation}
the gate's recall times exact match on the triggers it forwards. A stage-wise generation
score is Equation~\eqref{eq:e2e} with $g_i = 1$ for every trigger: an
\emph{oracle-gate ceiling} that no deployment realises. Because each $g_i(t)$ in
Equation~\eqref{eq:gate} can only switch off as $t$ rises, $\mathrm{EM}_{\mathrm{e2e}}$
cannot increase with the threshold; this is a property of the metric.

\subsection{Knowledge representation: the user memory profile}
\label{sec:knowledge}

The memory profile is the system's knowledge base: persistent, per wearer, accumulated
before the observed dialogue, and the generator's only information about the wearer
beyond the visible context. It ships with LlamaPIE's released dataset~\citep{chen2025llamapie},
one file per sample, and is used verbatim; its construction is outside our pipeline. A
profile is supplied whole and fixed per sample, so the corpus models the endpoint of a
memory that would, in deployment, accumulate across conversations.

\paragraph{Schema and segmentation.}
A profile has a \emph{profile} field of declarative prose and an \emph{events} map of
discrete episodes, both in prose rather than typed slots. For retrieval it is segmented
once, centrally, into \emph{facts} (one sentence of the profile field or of one event),
so every consumer indexes the same units; fragments below a minimum length are merged into
their predecessor, and events are ordered by a numeric-aware sort on their keys. A profile
holds \nFactsPerProfile{} facts on average across the \nRagTriggers{} test triggers.

\paragraph{Context window.}
At inference the profile is flattened --- profile field first, then events in key order
--- into one block that precedes the dialogue in the prompt. The dialogue is capped
separately at \nDialogueCap{} characters, which 60.4\% of training rows reach, so the
character cap rather than the sequence length binds. The profile adds \nPromptMemCost{}
tokens to every prompt (\nPromptWithMem{} against \nPromptNoMem{} on the test split). The
full prompt template is in Supplementary Section~\ref{sec:supp-genprompt}.

\paragraph{Containment.}
The target hint is present among the retrievable facts for \nContainment{}\% of triggers
and as a verbatim string in the profile for \nContainVerbatim{}\%. The first bounds what
any retrieval over the profile could supply, the second what pure extraction could: for
about half of all triggers the generator must produce the hint from the dialogue.

\subsection{Inference layers and decision logic}
\label{sec:inference}

The gating layer reads $D_{\leq i}$ and emits $p_i$ (Section~\ref{sec:gating} covers the
decoder and encoder gates measured). The threshold $t$ in Equation~\eqref{eq:gate} is an
operating parameter, specific to the model and the false-alarm budget, and is recalibrated
whenever the gate changes (Section~\ref{sec:implications}). When $g_i = 0$, processing at
that point ends with no output. When $g_i = 1$, the generation layer conditions on
$D_{\leq i}$ and $M_w$ and produces a single hint $h_i$ of one to three tokens; the length
constraint comes from the training corpus, not from the architecture (Supplementary
Figure~\ref{fig:hintlength}).

\subsection{Architecture}
\label{sec:architecture}

\begin{figure*}[width=\textwidth,pos=tp]
\centering
\includegraphics[width=\textwidth]{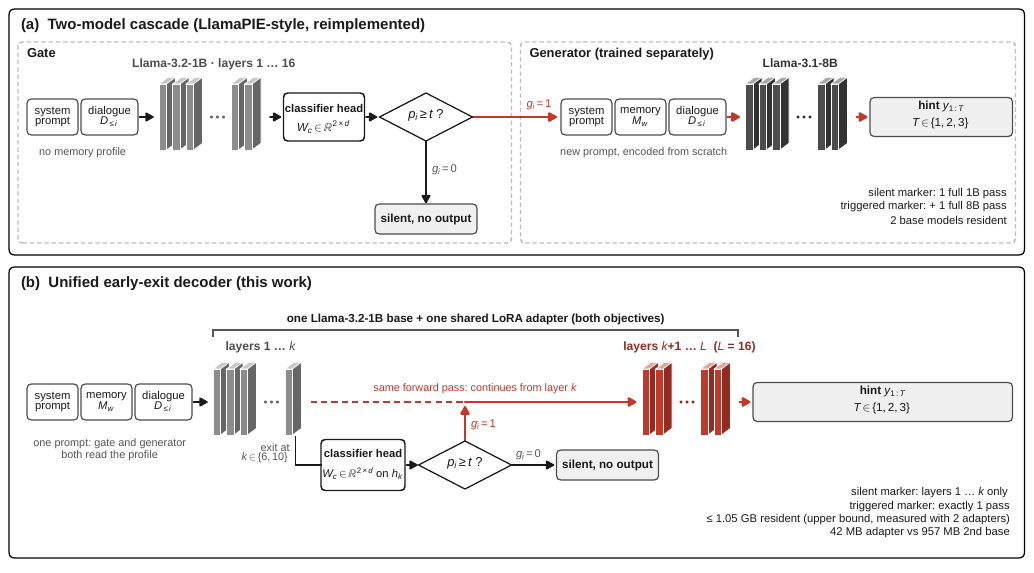}
\caption{The two realisations, for the Llama-3.2-1B / Llama-3.1-8B pairing. (a) Two-model
cascade: a separate gate reads the dialogue without the memory profile, and only the
generator adds it; a triggered marker pays two passes over two bases. (b) Unified decoder:
one base and one adapter read a single prompt that includes the profile; the head
$W_c \in \mathbb{R}^{2\times d}$ reads layer $k$, and a triggered marker continues through
layers $k{+}1..L$ in the same pass. The text below reports the control that gives the
cascade gate the profile too.}
\label{fig:architecture}
\end{figure*}

Figure~\ref{fig:architecture} shows the two realisations of the layers above. The
two-model cascade trains the gate and the generator as separate models, each paying its
own forward pass, and its gate reads the dialogue without the memory profile. The unified
decoder is one model reading one prompt that includes the profile: the gate head reads the
hidden state at layer $k$, and a triggered marker continues through layers $k{+}1..L$ in
the same forward pass. In full $L$-layer passes over a marker's prompt, before any hint
tokens are decoded, the two cost
\begin{equation}
c_i^{\mathrm{unified}} = \frac{k}{L} + g_i\left(1 - \frac{k}{L}\right),
\qquad
c_i^{\mathrm{cascade}} = 1 + g_i\, c_{\mathrm{gen}},
\label{eq:cost}
\end{equation}
where $c_{\mathrm{gen}}$ is one prefill of the separate generator in the same units. A
suppressed marker costs the unified model $k/L$ of a pass, the source of the ideal $L/k$
gate saving measured in Section~\ref{sec:cpucost}.

The unified gate also sees the memory profile, which the cascade gate does not, so we
measured what that information is worth. Retraining the two-stage 1B gate with the
profile prefixed to its input, recipe otherwise unchanged, moves its F1 from
\nMemGateBaseF{} to \nMemGateF{}: \nMemGateGain{} [\nMemGateGainLo{}, \nMemGateGainHi{}]
(95\%, clustered by dialogue), an interval that includes zero and lies within the noise
floor of Section~\ref{sec:gating}. The baseline here is that recipe's own retraining (the
focal-loss Llama-3.2-1B row of Supplementary Table~\ref{tab:loss}), not the reference gate
of Table~\ref{tab:gating}. Against the memory-aware gate, the unified gate at $k=10$ leads
by \nMemGateUTenLead{} F1 [\nMemGateUTenLeadLo{}, \nMemGateUTenLeadHi{}] and
\nMemGateUTenAurocDiff{} AUROC; the profile accounts for about \nMemGateShare{}\% of the
gap ([\nMemGateShareLo{}, \nMemGateShareHi{}]\%). Under a decision rule fixed before any
validation result existed, the $k=10$ advantage is therefore attributed to the
architecture. At $k=6$ the unified gate leads the memory-aware gate by \nMemGateUSixLead{}
F1 [\nMemGateUSixLeadLo{}, \nMemGateUSixLeadHi{}] and differs by \nMemGateUSixAurocDiff{}
AUROC, so a $k=6$ gate advantage may owe as much to the profile in its input as to the
architecture.

%% file: sections/03_task_protocol.tex
\section{Corpus and evaluation protocol}
\label{sec:task}

\subsection{Corpus and canonical marker index}
\label{sec:corpus}

All systems are evaluated on one canonical marker set of $N = \nMarkers{}$ test markers
(\nSilence{} silence, \nTriggers{} trigger, a \nTriggerRate{}\% trigger rate) over
\nDialogues{} dialogues, built without reference to any tokenizer and shared across
models. The dialogues come from three source corpora in LlamaPIE's release ---
\emph{perl}, \emph{soda} and \emph{synthetic} --- contributing \nPerlDialogues{},
\nSodaDialogues{} and \nSyntheticDialogues{} test dialogues. A marker's window is the
dialogue up to that point, truncated from the front at the \nDialogueCap{}-character cap,
so every marker is retained and coverage is 100\% for every model.

\subsection{Threshold selection}
\label{sec:convention}

Every gate operating point is selected by sweeping F1 over
$t \in \{0.01, 0.02, \ldots, 0.99\}$ on the \textbf{validation} split and reported on the
\textbf{test} split, using the same trained weights. Validation holds \nValMarkers{}
markers at \nValPrevalence{}\% trigger prevalence and is disjoint from test by dialogue.
Selecting on validation instead of test costs a mean of \nSelectionMeanDelta{} F1, and at
most \nSelectionWorstDelta{}, across \nSelectionSystems{} systems (for the unified gate,
\nProtocolBiasCorrectionKSix{} at $k=6$ and \nProtocolBiasCorrectionKTen{} at $k=10$); no
ranking changes.
The reimplemented two-stage gate's reportable figure is therefore \nTwoStageGateF{} at
$t = \nTwoStageThreshold{}$. Table~\ref{tab:gating} states the selection split for every
row, and AUPRC and AUROC, which do not depend on a threshold, accompany every operating
point.

\subsection{Metrics}
\label{sec:metrics}

\textbf{PA-F1} is the F1 of the gate's binary per-marker decision against the reference
trigger label, and \textbf{FAR} is the false-alarm rate, $\mathrm{FP}/(\mathrm{TN}+\mathrm{FP})$.
\textbf{End-to-end exact match} and \textbf{end-to-end precision} are
Equation~\eqref{eq:e2e}, computed over the \nTriggers{} test triggers and always reported
together.

Exact match is the primary automatic metric because it is the only one with a defensible
end-to-end form: a suppressed trigger has no string to score, and ``no hint delivered'' is
unambiguously not an exact match, whereas a BERTScore of zero would be an invented
convention. Exact match also counts a near-synonym as wrong, so we read it as a strict
lower bound on usefulness and report BERTScore beside it wherever generation is compared
stage-wise. Section~\ref{sec:evaluation} shows that graded metrics track human ratings
more closely.

BERTScore is computed with roberta-large, \texttt{idf=False} and
\texttt{rescale\_with\_baseline=False}, and rescaled figures are reported alongside, because
among the memory-only baselines, whose raw scores span only \nBaselineSpread{}, rescaling
moves every score by about \nBaselineShift{}, some \nRescaleVsSpread{} times that spread.

%% file: sections/04_gating.tex
\section{Gating}
\label{sec:gating}

\input{tables/tab01_gating}

Table~\ref{tab:gating} compares five decoder gates, one encoder under two training
recipes (RoBERTa-large), and two non-gate baselines under one protocol
(Section~\ref{sec:convention}). The best decoder is Llama-3.2-1B at \nTwoStageGateF{}
PA-F1, the gate the two-stage baseline uses throughout (\nTwoStageTestSelectedF{} if its
threshold were selected on test).

\paragraph{Noise floor.}
The reference band for every gate comparison is \nNoiseFloor{} F1, the difference between
two trainings of a nominally identical gate configuration. It comes from a single pair of
runs; the only other estimate of run-to-run spread is the three-seed standard deviation of
the unified gate (Table~\ref{tab:seeds}: \nSeedSdKSix{} at $k=6$, \nSeedSdKTen{} at
$k=10$). Later sections call this band the noise floor.

\paragraph{Encoder against decoders.}
RoBERTa-large under its own recipe reaches 0.8113 PA-F1, 1.11 points above the best
decoder ($p_\text{holm} = \nASLbeatsBestPHolm{}$, McNemar, corrected across all pairwise
comparisons in the table). A held-out audit over 200 splits puts the optimism of its
selected threshold at \nASLThresholdOptimismMean{} F1 on average and
\nASLThresholdOptimismHigh{} at the 95th percentile, comparable to the noise floor, so the
margin carries that caveat. Under the decoders' LoRA recipe the same architecture reaches
0.7483, below every decoder. Supplementary Section~\ref{sec:supp-gating} separates the
factors (Table~\ref{tab:equalisation}): context length contributes
$\nEqualisationContextDelta{}$ F1, the training recipe accounts for
$\nEqualisationRecipeDelta{}$ F1 of a same-architecture gap, and under a shared LoRA recipe
Llama-3.2-1B leads RoBERTa-large by $\nEqualisationArchDelta{}$ F1, a residual that mixes
architecture with an unequal training schedule. Plain cross-entropy is not beaten by any
alternative loss, and aggressive focal-loss settings reduce F1 (Table~\ref{tab:loss}).

\paragraph{Non-gate baselines.}
A logistic regression over 21 surface features --- turn position, sentence-final
punctuation and similar cues, no lexical content --- reaches 0.3456 PA-F1 (FAR 0.3108,
AUPRC 0.2368), and no single feature's point-biserial correlation with the label exceeds
0.1648 in magnitude: the decision is not recoverable from surface structure. A prompted
general-purpose LLM, on a stratified \nPromptedSubsampleN{}-marker test subsample, reaches
\nPromptedZeroShotF{} PA-F1 zero-shot and \nPromptedFewShotF{} with four worked examples,
with an AUROC of \nPromptedFewShotAUROC{} (prompts in Supplementary
Section~\ref{sec:supp-gateprompt}). What the fine-tuned gates learn is specific to this
corpus and is not supplied by instruction following.

%% file: tables/tab01_gating.tex
\begin{table*}[t]
\centering
\small
\setlength{\tabcolsep}{3.5pt}
\begin{tabular}{lccccc}
\toprule
System & $t$ & \textsc{source} & PA-F1 & FAR & AUPRC \\
\midrule
RoBERTa-large, ASL recipe & 0.59 & \textsc{val} & 0.8113 & 0.0304 & 0.8902 \\
\addlinespace[2pt]
Llama-3.2-1B & 0.60 & \textsc{val} & 0.8002 & 0.0351 & 0.8883 \\
Qwen2.5-0.5B & 0.60 & \textsc{val} & 0.7849 & 0.0386 & 0.8642 \\
SmolLM2-360M & 0.62 & \textsc{val} & 0.7800 & 0.0382 & 0.8596 \\
Gemma-3-1B & 0.24 & \textsc{val} & 0.7600 & 0.0359 & 0.8415 \\
Gemma-3-270M & 0.13 & \textsc{val} & 0.7212 & 0.0417 & 0.7784 \\
\addlinespace[2pt]
RoBERTa-large, matched LoRA & 0.70 & \textsc{val} & 0.7483 & 0.0536 & 0.8132 \\
\addlinespace[2pt]
Logistic regression, 21 structural features & 0.550 & \textsc{val} & 0.3456 & 0.3108 & 0.2368 \\
\addlinespace[2pt]
Prompted LLM gate, zero-shot & -- & \textsc{n/a} & 0.1381 & 0.0710 & 0.1543 \\
Prompted LLM gate, few-shot ($k=4$) & -- & \textsc{n/a} & 0.1914 & 0.1164 & 0.1545 \\
\bottomrule
\end{tabular}
\caption{Gate systems on the canonical marker index (\nMarkers{} markers, \nTriggers{} triggers), each at its own operating point; \textsc{source} says where that point was selected. Prompted-LLM rows use the model's own decision on a 2,000-marker Test subsample and are not directly comparable.}
\label{tab:gating}
\end{table*}

%% file: sections/05_generation.tex
\section{What the generator needs}
\label{sec:generation}

\input{tables/tab02_memory}

\paragraph{The memory profile.}
Table~\ref{tab:memory} scores the same adapter on the same \nTriggers{} triggers with the
profile present or absent at inference. Removing it costs \nMemoryEffectPrecise{}
exact-match points (McNemar $p < 0.001$); rescaled BERTScore falls from \nMemBertWith{} to
\nMemBertWithout{} and ROUGE-1 from \nMemRougeWith{} to \nMemRougeWithout{}. Three
controls attribute the effect to the profile's content. A mismatched profile, from another
sample, is indistinguishable from no profile ($p = \nMismatchVsNoneP{}$) and far below the
full profile ($p < 0.001$); decomposing the gain into a length and a content component
puts the length share at $\nMismatchLengthSharePct{}\%$. A model trained without any
profile performs close to the inference-time no-profile arm, so the effect does not depend
on train/inference matching; measured against that retrained model, the effect is
\nMemoryEffectRetrainedBaseline{} points. And the span-selection oracle, the best any extractive
strategy over the profile could do, reaches 0.9371 when the target is in the profile and
0 otherwise; the fine-tuned model scores below that ceiling even where extraction would
succeed, so it is not simply copying from the profile.

\paragraph{Generator scale.}
Within one family, Llama-3.1-8B and Llama-3.2-1B tie on exact match at the oracle-gate
stage ($p = \nScaleTieP{}$) and end to end ($p = \nScaleTieEndToEndP{}$;
Table~\ref{tab:unified}, ``1B gate + 1B gen'' and ``1B gate + 8B gen''). Across families,
Qwen2.5-7B reaches \nQwenCeiling{}\% oracle-gate exact match, ahead of both Llama
generators; it is an Instruct checkpoint compared with a base checkpoint, so family and
instruction tuning vary together in this comparison.

\begin{figure}[pos=tbp]
\centering
\includegraphics[width=\columnwidth]{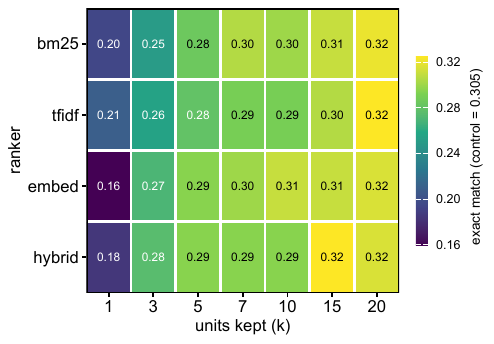}
\caption{Retrieval ablation: exact match for each ranker and depth on the 200-marker
ablation subsample, with memory view and prompt format at their control values. No cell
differs from the full-profile control after Holm correction across all 45
configurations.}
\label{fig:retrieval}
\end{figure}

\paragraph{Retrieval.}
Selecting a subset of the profile does not help. Exact match falls as the number of
retrieved units $k$ shrinks (Table~\ref{tab:memory}); $k=10$ and $k=5$ are significantly
below the full profile (Holm-corrected $p=0.0022$ and $p=0.0013$), and $k=20$, which keeps
87\% of the profile's units, is indistinguishable from it. Across ranker and depth
(Figure~\ref{fig:retrieval}), none of \nRagConfigs{} configurations beats the full
profile after Holm correction. The full profile already fits within the context budget, so
retrieval here only removes information.

%% file: tables/tab02_memory.tex
\begin{table}[t]
\centering
\small
\setlength{\tabcolsep}{4pt}
\begin{tabular}{>{\raggedright\arraybackslash}p{78pt}ccc}
\toprule
Memory condition & EM & BERTScore$_r$ & ROUGE-1 \\
\midrule
Full profile & 0.2947 & 0.5666 & 0.4757 \\
No profile at inference & 0.0474 & 0.3480 & 0.1258 \\
Mismatched profile & 0.0441 & 0.3384 & 0.1185 \\
Trained without profile & 0.0584 & 0.3812 & 0.1341 \\
Retrieval, $k=20$ & 0.2935 & -- & -- \\
Retrieval, $k=10$ & 0.2864 & -- & -- \\
Retrieval, $k=5$ & 0.2810 & -- & -- \\
Span-selection oracle & 0.4797 & -- & -- \\
\quad Full profile, target in profile & 0.9371 & -- & -- \\
\quad Full profile, target not in profile & 0.0000 & -- & -- \\
\bottomrule
\end{tabular}
\caption{What the generator needs: one adapter on the \nTriggers{} true triggers unless stated; BERTScore is rescaled. No retrieval configuration of \nRagConfigs{} beats the full profile after Holm correction.}
\label{tab:memory}
\end{table}

%% file: sections/06_unified.tex
\section{The unified early-exit model}
\label{sec:unified}

\input{tables/tab03_main}

The unified model shares one trunk and one LoRA adapter across both objectives
(Section~\ref{sec:architecture}). Table~\ref{tab:unified} is the single comparison
between it and the two-model pipelines; all differences are read against the noise floor
(Section~\ref{sec:gating}).

\begin{figure}[pos=tbp]
\centering
\includegraphics[width=\columnwidth]{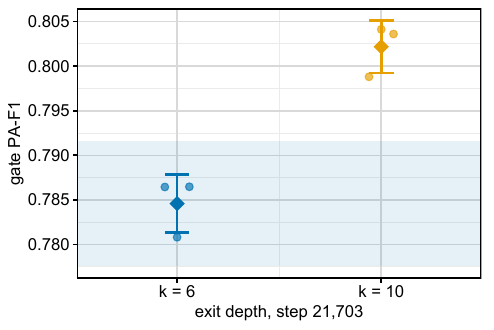}
\caption{Gate PA-F1 at the two exit depths, step 21{,}703: seeds as points, mean
$\pm$1 SD as a diamond, and the noise floor shaded around the $k=6$ mean. The $k{=}10$
minus $k{=}6$ gap exceeds the noise floor.}
\label{fig:exitdepth}
\end{figure}

\paragraph{Gate quality.}
At the final checkpoint (step \nFinalStep{}, one seed per depth), the unified gate,
validation-selected and reported on test like every other gate, reaches
\nUnifiedGateFKSix{} PA-F1 at $k=6$ and \nUnifiedGateFKTen{} at $k=10$, against
\nTwoStageGateF{} for the two-stage gate. The $k=6$ margin (\nUnifiedGateGapKSix{}) is
within the noise floor; the $k=10$ margin (\nUnifiedGateGapKTen{}) exceeds it. The
three-seed replication at step \nSeedStep{} (Table~\ref{tab:seeds},
Figure~\ref{fig:exitdepth}) refines this by depth: the $k=10$ mean, \nSeedMeanKTen{}
($\pm$\nSeedSdKTen{}), ties the two-stage gate (\nSeedGapKTen{}), and the $k=6$ mean,
\nSeedMeanKSix{} ($\pm$\nSeedSdKSix{}), is \nSeedGapKSix{} below it, outside the noise
floor. Multi-seed evidence therefore supports gate parity at $k=10$; at $k=6$ parity rests
on the single final-checkpoint seed. Seed dispersion is measured at step \nSeedStep{}
only, since pooling it with a point estimate from another step would misstate it.

\paragraph{Exit depth.}
$k=10$ leads $k=6$ in gate F1 by \nGateGapEarly{} at step \nSeedStep{} and by
\nGateGapLate{} at the final checkpoint, both above the noise floor; the gap did not grow
with further training, so it appears fixed by where the decision is read. End to end,
the two depths are within the noise floor of each other in exact match
(\nUnifiedEndToEndEmKSix{} at $k=6$, \nUnifiedEndToEndEmKTen{} at $k=10$;
Table~\ref{tab:unified}), while $k=6$ costs less wall clock (Figure~\ref{fig:compute}).
Exit depth is therefore mainly a cost decision.

\begin{figure}[pos=tbp]
\centering
\includegraphics[width=\columnwidth]{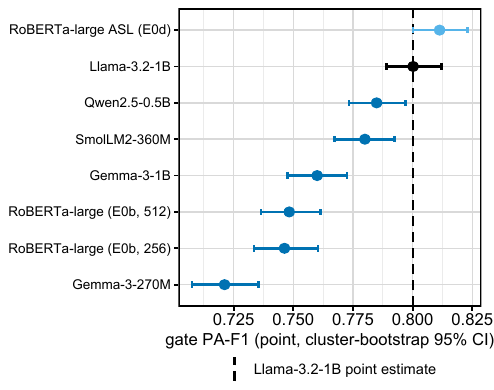}
\caption{Gate PA-F1 with cluster-bootstrap 95\% CIs for every gate. The dashed line is
the two-stage Llama-3.2-1B gate; darker markers have CIs that exclude it.}
\label{fig:forest}
\end{figure}

%% file: tables/tab03_main.tex
\begin{table*}[t]
\centering
\small
\setlength{\tabcolsep}{4pt}
\begin{tabular}{>{\raggedright\arraybackslash}p{125pt}cccccccc}
\toprule
System & Gate F1 & FAR & Gen.\ EM & BERTScore & E2E EM & E2E prec. & Wall (h) & Models \\
\midrule
Unified, exit at layer $k=6$ & 0.8022 & 0.0359 & 0.3299 & 0.9288 & 0.2929 & 0.2665 & 0.748 & 1 / 1 \\
Unified, exit at layer $k=10$ & 0.8164 & 0.0292 & 0.3218 & 0.9269 & 0.2908 & 0.2761 & 1.005 & 1 / 1 \\
\addlinespace[2pt]
Unified, gate head on final layer $k=16$ (step 21,703, gate-only) & 0.7975 & 0.0324 & n/a & n/a & n/a & n/a & n/a & 1 / 1 \\
\midrule
Two-model, 1B gate + 1B gen & 0.8002 & 0.0351 & 0.2971 & 0.9237 & 0.2619 & 0.2387 & 1.318 & 2 / 2 \\
Two-model, 1B gate + 8B gen & 0.8002 & 0.0351 & 0.2947 & 0.9269 & 0.2548 & 0.2322 & 2.551 & 2 / 2 \\
\bottomrule
\end{tabular}
\caption{The unified model against the two-model pipelines at the final checkpoint (step 53,772, one seed per depth). E2E EM and E2E prec.\ are Equation~\eqref{eq:e2e}; wall clock is early-exit-only. The final-layer ($k=16$) row is a gate-only control from step 21,703.}
\label{tab:unified}
\end{table*}

\begin{table}[t]
\centering
\small
\begin{tabular}{>{\raggedright\arraybackslash}p{74pt}ccc}
\toprule
Configuration & Seeds & Mean gate F1 & SD \\
\midrule
Unified $k=6$, step 21,703 & 3\,/\,3 & 0.7846 & 0.0033 \\
Unified $k=10$, step 21,703 & 3\,/\,3 & 0.8022 & 0.0029 \\
\bottomrule
\end{tabular}
\caption{Run-to-run dispersion, reported at the step each seed was measured at. Values from different steps are never pooled.}
\label{tab:seeds}
\end{table}

%% file: sections/07_endtoend.tex
\section{End to end and cost}
\label{sec:endtoend}

\begin{figure}[pos=tbp]
\centering
\includegraphics[width=\columnwidth]{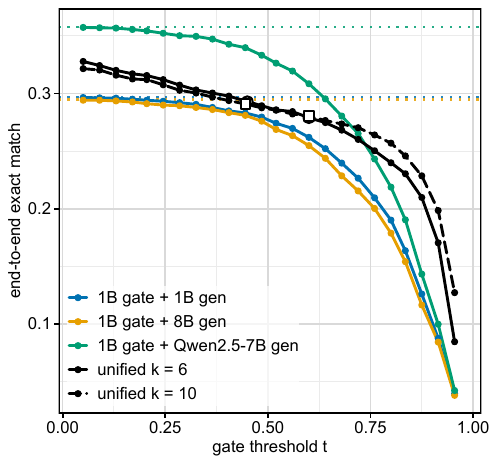}
\caption{End-to-end exact match against gate threshold for every system; dotted lines are
the oracle-gate ceilings. The curves fall by construction (Equation~\eqref{eq:e2e}), so
they show how much of each ceiling a threshold gives up. Markers on the unified curves
are the gate-F1-optimal and $t=0.60$ operating points.}
\label{fig:monotonicity}
\end{figure}

\paragraph{From ceiling to deployment.}
Every stage-wise generation score in Section~\ref{sec:generation} is an oracle-gate
ceiling (Section~\ref{sec:problem}). Behind the two-stage gate at its deployed operating
point (precision \nTwoStagePrecision{}, recall \nTwoStageRecall{}), end-to-end exact match
is \nEndToEndExactMatch{} against a ceiling of \nOracleCeilingOneB{} for the 1B generator,
\nEndToEndExactMatchEightB{} against \nOracleCeilingEightB{} for the 8B, and
\nEndToEndExactMatchQwen{} against \nOracleCeilingQwen{} for Qwen2.5-7B.
Figure~\ref{fig:monotonicity} plots end-to-end exact match over \nSweepPoints{}
thresholds. Because it cannot rise with the threshold, it is never read alone: end-to-end
precision, which charges false alarms, is \nEndToEndPrecision{} at the deployed point, and
tightening to a 1\% false-alarm budget raises it to \nEndToEndPrecisionOnePct{} while
end-to-end exact match falls to \nEndToEndExactMatchOnePct{}; Section~\ref{sec:implications}
discusses the choice of budget. Non-matching hints split across partial overlaps,
grounded but incorrect answers, and hallucinations in similar shares for every generator
(Table~\ref{tab:errors}).

\input{tables/tab13_errors}

\paragraph{Cost, and the basis it is measured on.}
At the deployed operating point the gate cuts generator invocations by
\nInvocationCut{}$\times$. That cut is worth \nSpeedupEightB{}$\times$ in wall clock
against an 8B generator and \nSpeedupOneB{}$\times$ against a 1B one, because the gate
itself takes \nGateShare{}\% of the gated 1B pipeline's runtime on our per-marker harness
(about 53\% once the measured prefix-cache ratio is applied). We report wall clock on an
early-exit-only basis, the only optimisation measured for both architectures. Prefix
caching applies to both, but the two-model pipeline's prompt format is the more cacheable
(80.3\% of tokens against 58.1\%) and its cached rate is not measured, because its gate
adapter was not retained; combined-optimisation figures are therefore an upper bound on
the unified model's advantage and never the headline.

\begin{figure}[pos=tbp]
\centering
\includegraphics[width=\columnwidth]{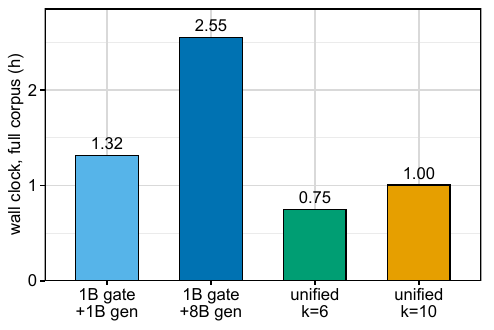}
\caption{Wall clock over all \nMarkers{} markers (early-exit-only basis) for the two
two-model pipelines and the unified model at both exit depths.}
\label{fig:compute}
\end{figure}

\paragraph{The unified model end to end.}
On this basis, at its final checkpoint, the unified model reaches higher end-to-end exact
match and end-to-end precision than both two-model pipelines at both depths
(Table~\ref{tab:unified}), with \nSpeedupKSix{}$\times$ less wall clock than the matched
1B pipeline at $k=6$ and \nSpeedupEightBKSix{}$\times$ less than the 8B one
(Figure~\ref{fig:compute}). This is a result on automatic measures at one checkpoint;
Section~\ref{sec:evaluation} reports how people rate the same hints.

%% file: tables/tab13_errors.tex
\begin{table*}[t]
\centering
\small
\setlength{\tabcolsep}{4pt}
\begin{tabular}{lccccc}
\toprule
Generator & Exact & Partial & Grounded, & Halluci- & Halluc. \\
 & & overlap & other & nated & share \\
\midrule
Llama-3.2-1B (1B gen) & 29.7 & 17.8 & 35.9 & 16.6 & 23.6 \\
Llama-3.1-8B (8B gen) & 29.5 & 27.9 & 23.5 & 19.1 & 27.1 \\
Qwen2.5-7B (7B gen) & 35.8 & 18.5 & 27.3 & 18.4 & 28.6 \\
\bottomrule
\end{tabular}
\caption{Error types per generator, as percentages of the \nTriggers{} true triggers; \emph{hallucinated share} is hallucinations as a fraction of non-exact responses.}
\label{tab:errors}
\end{table*}

%% file: sections/08_evaluation.tex
\section{Human evaluation and evaluation validity}
\label{sec:evaluation}

\input{tables/tab04_human}

\paragraph{Design.}
The human study followed a pre-registered analysis plan, time-stamped and hash-pinned in
version control before any rating existed. It sampled \nHumanDialoguesPlanned{} test
dialogues containing \nHumanWhispersPlanned{} hints, each rendered in
\nHumanRenderings{} versions and assigned to \nHumanAnnotatorsPlanned{} annotators, who
rated every hint on LlamaPIE's five-point uptake scale with the subsequent turns visible
(instructions in Supplementary Section~\ref{sec:supp-annotators}). Every rendering places
its hints at the same gold trigger markers, so the study compares what each system says,
not when it speaks.

\paragraph{As run.}
The study deviated from the plan three times. One annotator's ratings were scored,
unblinded, during annotation; annotation was then stopped early, after that interim look;
and missing ratings were excluded per item rather than per hint, a scoring change made
after the data were seen, which retained the second annotator's uptake ratings. The
analysed data are \nHumanRatings{} uptake ratings over \nHumanDialogues{} of the
\nHumanDialoguesPlanned{} dialogues: \nHumanAnnOne{} of one annotator's \nHumanAnnOneOf{}
assigned hints, \nHumanAnnFive{} from a second annotator on \nHumanAnnFiveConvs{}
conversations, and \nHumanAnnThree{} from a third. On the \nHumanAlphaUnits{} hints both
annotators rated, Krippendorff's ordinal $\alpha = \nHumanAlpha{}$; identical texts shown
twice under different codes received the same rating on \nHumanRetestOneB{}\% and
\nHumanRetestEightB{}\% of pairs.

\paragraph{Memory.}
With the profile, the 8B generator's hints average \nHumanHoneMem{} in uptake against
\nHumanHoneNoMem{} without it, a difference of \nHumanHoneDiff{} [\nHumanHoneLo{},
\nHumanHoneHi{}] (95\%, clustered by dialogue). The primary outcome agrees with the
automatic metrics: the profile makes hints more useful to the person hearing them.

\paragraph{Unified model.}
Against the matched 1B two-stage pipeline the unified model is level at both depths:
\nHumanUSixOneBDiff{} [\nHumanUSixOneBLo{}, \nHumanUSixOneBHi{}] at $k=6$ and
\nHumanUTenOneBDiff{} [\nHumanUTenOneBLo{}, \nHumanUTenOneBHi{}] at $k=10$ (98.75\%,
Bonferroni over the four non-inferiority tests). Both intervals extend below
$-$\nHumanMargin{}, so non-inferiority at the planned margin is not shown. Against the 1B+8B
pipeline it is lower, by \nHumanUSixEightBDiff{} [\nHumanUSixEightBLo{},
\nHumanUSixEightBHi{}] and \nHumanUTenEightBDiff{} [\nHumanUTenEightBLo{},
\nHumanUTenEightBHi{}], although it is higher on end-to-end exact match
(Table~\ref{tab:unified}); the separate-arm tests agree. As an exploratory, unplanned
comparison, the 8B generator rates \nHumanEightOverOneB{} points above the 1B one
(\nHumanHoneMem{} against \nHumanOneBMean{}), whereas the two tie on exact match.

\paragraph{Which automatic signal to use.}
The native judge of LlamaPIE, where computed, tracks human uptake most closely
($r = \nHumanCorrNativeEight{}$ on the 8B renderings, $\nHumanCorrNativeOne{}$ on the 1B),
followed by ROUGE-1 ($\nHumanCorrRougeEight{}$, $\nHumanCorrRougeOne{}$) and BERTScore
($\nHumanCorrBertEight{}$, $\nHumanCorrBertOne{}$), then exact match
($\nHumanCorrEmEight{}$, $\nHumanCorrEmOne{}$; Table~\ref{tab:human}b). We therefore
recommend the native judge as the automatic judge for this task, a choice made after this
comparison. The ranking rests on two renderings: the native judge scored only the 8B and
1B ones, it was not run on the no-memory output, and per-item BERTScore and ROUGE-1 were
not computed for the unified model. Under the native prompt, Qwen2.5-7B scores 3.83--4.07
per corpus, against 3.94--4.21 reported by LlamaPIE under the same prompt. Our
reconstruction of LlamaPIE's rubric from its published definitions agrees poorly with the
native judge and is not used for any claim (Supplementary Section~\ref{sec:supp-judgeagree}).
The item asking whether a hint should have been withheld received \nHumanSilenceAnswers{}
answers, almost all ``no'', and supports no claim.

%% file: tables/tab04_human.tex
\begin{table*}[t]
\centering
\small
\setlength{\tabcolsep}{4pt}
\begin{tabular}{llrrrlrl}
\toprule
\multicolumn{8}{l}{\textit{(a) Pre-registered contrasts, mean uptake (1--5)}} \\
Test & A vs B & A & B & A$-$B & CI & level & reading \\
\midrule
H1 (primary) & 8B, memory vs 8B, no memory & 4.01 & 3.18 & +0.83 & [+0.48, +1.23] & 95\% & higher \\
\addlinespace[3pt]
H2a & Unified $k{=}6$ vs 1B & 3.64 & 3.60 & +0.04 & [$-$0.31, +0.41] & 98.75\% & NI not shown \\
H2b & Unified $k{=}10$ vs 1B & 3.58 & 3.60 & $-$0.02 & [$-$0.35, +0.30] & 98.75\% & NI not shown \\
\addlinespace[3pt]
H2a & Unified $k{=}6$ vs 8B, memory & 3.64 & 4.01 & $-$0.38 & [$-$0.69, $-$0.04] & 98.75\% & lower; NI not shown \\
H2b & Unified $k{=}10$ vs 8B, memory & 3.58 & 4.01 & $-$0.44 & [$-$0.83, $-$0.07] & 98.75\% & lower; NI not shown \\
\addlinespace[3pt]
H2b, separate arm & Unified $k{=}10$ vs 1B (copy) & 3.58 & 3.64 & $-$0.06 & [$-$0.41, +0.27] & 98.75\% & NI not shown \\
H2b, separate arm & Unified $k{=}10$ vs 8B, memory (copy) & 3.53 & 3.98 & $-$0.45 & [$-$0.84, $-$0.08] & 98.75\% & lower; NI not shown \\
\bottomrule
\end{tabular}
\par\medskip
\begin{tabular}{lrrrrrr}
\toprule
\multicolumn{7}{l}{\textit{(b) Correlation of human uptake with automatic signals}} \\
Rendering & EM & BERTScore & ROUGE-1 & judge, ref.-based & judge, reconstr. & judge, native \\
\midrule
8B, memory & 0.31 & 0.47 & 0.51 & 0.45$^*$ & 0.28$^*$ & 0.55 \\
8B, no memory & 0.30 & 0.37 & 0.44 & 0.51$^*$ & 0.27$^*$ & --- \\
Unified $k{=}6$ & 0.43 & --- & --- & 0.54$^*$ & 0.29$^*$ & --- \\
Unified $k{=}10$ & 0.44 & --- & --- & 0.51$^*$ & 0.17$^*$ & --- \\
1B & 0.43 & 0.61 & 0.62 & 0.55$^*$ & 0.44$^*$ & 0.67 \\
\bottomrule
\end{tabular}
\caption{Human evaluation. \emph{(a)} Pre-registered contrasts (labels as in the study's pre-registration): mean uptake (1--5) per arm, paired difference A$-$B and dialogue-clustered CI. H2 tests are Bonferroni-adjusted, non-inferiority margin 0.25. \emph{(b)} Pearson $r$ between human uptake and each automatic signal over 136 hints; $^*$\,$n < 60$. 1,144 ratings over 32 of 100 dialogues, mostly one rater (Section~\ref{sec:evaluation}).}
\label{tab:human}
\end{table*}

%% file: sections/11_implications.tex
\section{Practical implications}
\label{sec:implications}

\subsection{Recalibrate the threshold whenever the gate model changes}

A threshold carried across a gate swap changes the false-alarm rate by an amount and in a
direction the models' headline scores do not predict. The threshold that attains a 2\%
false-alarm budget ranges by more than a factor of three across the six deployment
candidates, and in the worst case \nThresholdCollapseModel{} loses \nThresholdCollapse{}
F1 points when run at another model's operating point (Supplementary
Table~\ref{tab:operating}). Per-model calibration on held-out data is a deployment
requirement.

\subsection{Select on calibration when thresholds adapt}

A system that adapts its threshold from user feedback consumes the gate's probabilities,
not only its ranking. The best-ranking gate is also the worst-calibrated, with an expected
calibration error several times that of the best-calibrated model (Supplementary
Table~\ref{tab:calibration} and Figure~\ref{fig:reliability}), so an integrator building
feedback adaptation should select on calibration, or calibrate post hoc and re-measure.

\subsection{Choose the false-alarm budget deliberately}

Because a false alarm and a miss cost the wearer different things
(Section~\ref{sec:problem}), the false-alarm budget is a deployment decision. At the
deployed operating point the system emits \nEmitted{} hints, \nSpuriousShare{}\% of them
where the corpus says it should have stayed silent. Tightening the budget trades
end-to-end exact match for end-to-end precision (Section~\ref{sec:endtoend}): fewer wrong
interruptions, bought with fewer right ones.

\subsection{Resident weight cost}

Serving both decisions from one base saves memory whenever gate and generator share an
architecture. A probe holding one 4-bit 1B base with a gate adapter, a generator adapter
and the classifier head resident uses \nResidentWeights{}\,GB of weights: the second
adapter costs \nSecondAdapterMB{}\,MB where a second base would cost \nSecondBaseMB{}\,MB,
a saving of \nSharedTrunkSavingMB{}\,MB. The unified model needs one adapter, so this
figure is an upper bound on its weights, whose own footprint was not measured separately.

\subsection{Cost on a CPU}
\label{sec:cpucost}

The unified gate runs on a CPU; a graph runtime turns its early exit into time saved, and
4-bit weights save memory rather than time. We timed both final checkpoints on a 4-core
cloud CPU at batch size one, in every runtime and weight format (Supplementary
Table~\ref{tab:cpulatency}). In ONNX Runtime with NF4 weights, the format the model was
trained in, the gate costs \nCpuGateNfFourKSix{}\,ms per marker at $k=6$ and
\nCpuGateNfFourKTen{}\,ms at $k=10$, and a triggered marker costs a further
\nCpuHintNfFourKSix{}\,s and \nCpuHintNfFourKTen{}\,s to generate its hint. This variant
reproduces the GPU model: gate probabilities correlate at $r = \nCpuAgreeNfFourKSix{}$ and
$\nCpuAgreeNfFourKTen{}$, decisions agree on \nCpuDecisionNfFourKSix{}\% and
\nCpuDecisionNfFourKTen{}\% of sampled markers, and hints are identical on
\nCpuHintMatchNfFourKSix{}\% and \nCpuHintMatchNfFourKTen{}\% of \nCpuHintMarkers{}
triggered markers. Each marker is timed as an independent forward pass over its full
prompt; cross-marker caching was not measured.

In ONNX Runtime the early exit delivers a saving consistent with the ideal $16/k$ of
Equation~\eqref{eq:cost} ($\nCpuSpeedupNfFourKSix{}\times$ at $k=6$ and
$\nCpuSpeedupNfFourKTen{}\times$ at $k=10$ in NF4), against
$\nCpuSpeedupPtFullKSix{}\times$ and $\nCpuSpeedupPtFullKTen{}\times$ for PyTorch in eager
mode. NF4 does not speed the gate up at batch size one (\nCpuGateNfFourKSix{} against
\nCpuGateOrtFullKSix{}\,ms in fp32) but reduces memory (\nCpuDiskNfFour{}\,GB against
\nCpuDiskFull{}\,GB); half precision is slower than fp32 on these CPUs, which lack the
instructions that make it fast (Supplementary Section~\ref{sec:supp-cpu}).

%% file: sections/12_threats.tex
\section{Limitations}
\label{sec:threats}

Every headline result rests on one primary corpus whose dialogues and trigger labels were
generated by a language model, so its labels mark moments a model judged a hint would
help, not moments a wearer needed one. On a naturally recorded meeting corpus, a
fine-tuned gate performs on par with an observable-metadata baseline (AUROC
\nTransferAurocFineTuned{} against \nTransferAurocBaseline{}; Supplementary
Section~\ref{sec:transfer}). Most configurations were trained once, so our
confidence intervals are sampling intervals over fixed models, and differences below the
noise floor (Section~\ref{sec:gating}) are not resolved. Training budgets differ: the
unified model ran \nUnifiedEpochs{} epochs (\nFinalStep{} steps) over marker-level rows and
the two-stage gate \nTwoStageEpochs{} epochs over dialogue-level samples, so comparisons
are between systems as trained. A second-family unified model, further epochs and other
exit depths were not run. The human evaluation rests on \nHumanRatings{} ratings, mostly
from one rater (Section~\ref{sec:evaluation}). The two-stage gate's cost under prefix
caching was not measured (Section~\ref{sec:endtoend}). No consumer device was measured,
and the system is evaluated on transcripts, so speech-recognition errors are absent from
every number.

%% file: sections/10_conclusion.tex
\section{Conclusion}
\label{sec:conclusion}

Scoring a proactive assistant's two decisions together changes what the evidence shows.
Stage-wise generation scores are oracle-gate ceilings; because end-to-end exact match
cannot rise as the gate tightens, a gate is justified by the precision it adds and by the
compute it saves, and that saving depends on the size of the generator. Hint quality
comes from the persistent memory profile rather than from generator scale or retrieval,
and a human evaluation supports the memory effect.

One compact decoder that gates from an intermediate layer and generates in the same pass
matches or exceeds the two-model pipeline on the automatic end-to-end measures at lower
wall-clock cost (Table~\ref{tab:unified}). Across three seeds its $k=10$ gate ties the
two-stage reference gate (Section~\ref{sec:unified}); at the final checkpoint it leads a
two-stage gate that also reads the memory profile, so that lead is attributable to the
architecture (Section~\ref{sec:architecture}). People rate its
hints level with the matched 1B pipeline and below the 8B one
(Section~\ref{sec:evaluation}). For evaluating such systems, LlamaPIE's native judge and graded
metrics track human ratings more closely than exact match, which remains the one metric
with an end-to-end form.

%% file: frontmatter/declarations.tex
\section*{Ethical considerations}
\label{sec:ethics}

The system listens continuously to a conversation on behalf of one wearer, in the
presence of people who did not consent to being monitored and cannot see or query the
profile it keeps. That profile is persistent personal data about the wearer, accumulated
across conversations, and carries the risks of any long-lived personal record: it is a
target for compromise, and its content --- names, relationships, recurring topics --- is
information a wearer may not want kept indefinitely. A false alarm interrupts the
conversation for everyone present, so bystanders who never opted in also bear the cost
of the system's errors. Non-user consent, disclosure norms and a retention policy for the
profile are deployment requirements this work does not address.

ICSI participants (Supplementary Section~\ref{sec:transfer}) consented to being recorded for meeting
research, not to a proactive-assistance study centred on one participant, in which the
other participants become sources of the wearer's triggers. We use the corpus only as
released, under its own licence. One artefact of this repurposing was found and
corrected: ICSI's digit-reading task produced number strings that an entity extractor
mistook for dates, creating spurious profile content and triggers, so number-word terms
are excluded at extraction.

Models supervise and judge this work at two points. The primary corpus's trigger labels
were generated by a language model reading synthetic dialogue, which is why
Supplementary Section~\ref{sec:transfer} tests a behaviourally labelled corpus; and the LLM judges are
themselves models, whose limits Supplementary Section~\ref{sec:supp-judgeagree} reports.
We treat both as properties of the evidence that bound what it can support.

\section*{Data availability}
\label{sec:data}

We release the construction and validation code rather than the derived corpus itself,
together with every artifact the paper's numbers are computed from: the canonical marker
index, per-marker gate predictions for every system, generation outputs, all judge
prompts and all raw judgments, the evaluation and scoring scripts, and the run manifests.
Every figure and table in this paper is regenerated from those artifacts by the released
scripts, so each reported number is reproducible without the corpus being redistributed.

The dialogue corpus itself is \emph{not} redistributed. It derives from three upstream
resources whose licences differ: one restricts use to non-commercial research, and one
states no dataset licence at all. Rather than pass on material we may not be entitled to
pass on, we release the pipeline that rebuilds the derived corpus from those upstream
releases, which a reader obtains separately under the upstream projects' own terms. The
construction is deterministic, and the released canonical index lets a reader verify that
their rebuild matches ours marker for marker.

\ifanonymized
Code and artifacts will be made available on publication. No link is given here, in
keeping with anonymous review.
\else
Code and artifacts will be made available on publication, and a link will be added to
this paper when the release is public.
\fi

\section*{Declaration of competing interests}
\label{sec:coi}

The authors declare that they have no known competing financial interests or personal
relationships that could have appeared to influence the work reported in this paper.

\section*{Declaration of generative AI and AI-assisted technologies in the manuscript preparation process}
\label{sec:genai}

During the preparation of this work the authors used Claude (Anthropic) to draft and
revise text and to check cited numbers against the experimental outputs. After using
this tool, the authors reviewed and edited the content as needed and take full
responsibility for the content of the published article.

%% file: frontmatter/acknowledgements.tex
\section*{Acknowledgements}
\label{sec:acknowledgements}

We thank the authors of LlamaPIE for releasing the code and data on which this work's
corpus is built, and the creators of the SODA, PerLTQA and ICSI corpora, whose
public releases it draws on. We thank Kaggle for the free GPU quota on which our models
were trained and evaluated. We are grateful to the annotators who gave their time to the
human evaluation.

%% file: frontmatter/funding.tex
This work received no specific grant from any funding agency in the public,
commercial, or not-for-profit sectors. Model training and evaluation were carried out
on Kaggle's free-tier notebook GPU allocation, and the CPU latency measurements on
Modal cloud CPU containers; no grant or institutional compute allocation was used. The
primary corpus's construction pipeline builds on publicly released resources from the
LlamaPIE project (Chen, Batchelder, Liu, Smith and Gollakota, 2025), used under that
project's own release terms rather than under any funded data-acquisition arrangement.

%% file: supplement/body.tex
\providecommand{\suppfigwidth}{\columnwidth}

\setcounter{section}{0}
\renewcommand{\thesection}{S\arabic{section}}
\setcounter{table}{0}
\renewcommand{\thetable}{S\arabic{table}}
\setcounter{figure}{0}
\renewcommand{\thefigure}{S\arabic{figure}}

\section{Gate training protocol: what explains the encoder's lead}
\label{sec:supp-gating}

The best encoder gate (RoBERTa-large, 0.8113 PA-F1) beats the best decoder gate, but the
same architecture trained under the decoders' own LoRA recipe falls to 0.7483
(Section~\ref{sec:gating}). Table~\ref{tab:equalisation} separates the factors, using
rows that differ in one respect at a time. Holding architecture and recipe fixed,
context length explains none of the gap: quadrupling RoBERTa's window from 256 to 512
tokens moves PA-F1 by only $\nEqualisationContextDelta{}$, inside the noise floor, and
the paired accuracy test, if anything, favours the \emph{shorter} window
($p_\text{holm} = \nEqualisationContextPHolm{}$). Holding architecture and context fixed
instead, the recipe explains almost all of a same-architecture gap: the ASL full
fine-tuning recipe beats matched LoRA by $\nEqualisationRecipeDelta{}$ F1 at identical
context and epoch count ($p_\text{holm} < 0.001$). What is left is
the decoder-vs-encoder gap under a \emph{shared} LoRA recipe: Llama-3.2-1B still leads
RoBERTa-large by $\nEqualisationArchDelta{}$ F1
($p_\text{holm} < 0.001$). Epoch count is not equalised in that pair
(1 against 4), so this residual is architecture and training schedule together, not
architecture alone.

\input{tables/tab05_equalisation}

The loss function is not the mechanism (Table~\ref{tab:loss}). For both models where it
was ablated, no alternative to plain cross-entropy --- class-weighted cross-entropy, or
focal loss at $\gamma = 1.5$ --- reaches significance against it (closest
$p_\text{holm} = \nLossAblationWorstPHolm{}$). Sweeping the focal-loss hyperparameters on
Qwen2.5-0.5B, $\gamma \in \{0, 1, 1.5\}$ are statistically indistinguishable, but
$\gamma = 2$ costs $\nFocalGammaTwoDelta{}$ F1 ($p_\text{holm} < 0.001$)
and raising the trigger-class weight $\alpha$ from 1 to 4 costs $\nFocalAlphaFourDelta{}$
($p_\text{holm} < 0.001$). Moderate reweighting is harmless, aggressive
reweighting hurts, and neither explains the ASL recipe's lead, which pairs focal loss
with full fine-tuning rather than with LoRA.

\input{tables/tab06_loss}

\section{Thresholds do not transfer between gate models}
\label{sec:supp-threshold}

Fixing a false-alarm budget and asking which threshold attains it, the answer ranges by
more than a factor of three across the six deployment candidates at a 2\% budget, and
further once the protocol-equalisation controls of Section~\ref{sec:supp-gating} are
included (Table~\ref{tab:operating}). Two models a few points of PA-F1 apart need
operating points at opposite ends of the probability scale. In the worst case,
\nThresholdCollapseModel{} loses \nThresholdCollapse{} F1 points when run at another
model's operating point.

\input{tables/tab07_operating}

\section{Ranking quality and calibration}
\label{sec:supp-calibration}

\begin{figure}[pos=tbp]
\centering
\includegraphics[width=\suppfigwidth]{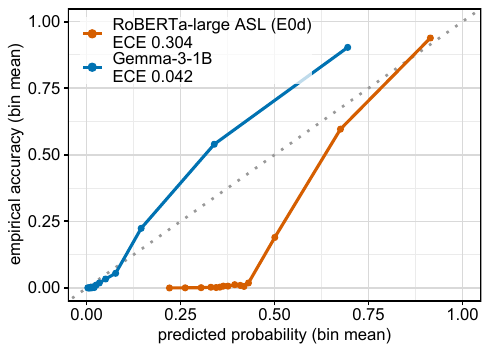}
\caption{Reliability diagram for the best-ranking gate (RoBERTa-large ASL) and the
best-calibrated one (Gemma-3-1B). The dotted diagonal is perfect calibration.}
\label{fig:reliability}
\end{figure}

Table~\ref{tab:calibration} measures ranking (AUPRC) and probability quality (ECE, MCE,
Brier) on identical markers. The best-ranking gate is also the worst-calibrated, with an
expected calibration error several times that of the best-calibrated model, which ranks
lower (Figure~\ref{fig:reliability}).

\input{tables/tab08_calibration}

\section{Hint length and format compliance}
\label{sec:supp-format}

\begin{figure}[pos=tbp]
\centering
\includegraphics[width=\suppfigwidth]{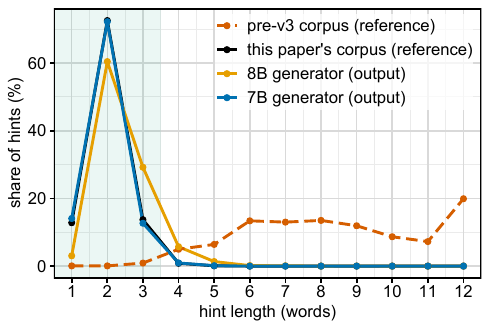}
\caption{Hint length in words. The earlier corpus (\emph{v1}) put 99.0\% of its
reference hints outside the one-to-three-word window; this paper's corpus (\emph{v3})
puts 99.1\% inside it, and both generators stay close (92.7\% and 99.0\%).}
\label{fig:hintlength}
\end{figure}

Figure~\ref{fig:hintlength} shows the length constraint as a distribution. Length
compliance is not a quality measure: the no-memory ablation complies \emph{more} often
than the memory-bearing system while being far worse on every content metric
(Table~\ref{tab:memory}, Section~\ref{sec:generation}).

\section{CPU cost for every runtime and weight format}
\label{sec:supp-cpu}

Table~\ref{tab:cpulatency} gives every variant behind Section~\ref{sec:cpucost}.
Exiting after $k$ of 16 layers can at best cost $k/16$ of a full pass
(Equation~\eqref{eq:cost}), a $\nCpuIdealSpeedupKSix{}\times$ saving at $k=6$ and
$\nCpuIdealSpeedupKTen{}\times$ at $k=10$. ONNX Runtime's within-container ratios are
consistent with that ceiling --- $\nCpuSpeedupOrtFullKSix{}\times$ and
$\nCpuSpeedupOrtFullKTen{}\times$ at fp32, $\nCpuSpeedupNfFourKSix{}\times$ and
$\nCpuSpeedupNfFourKTen{}\times$ at NF4, the $k=6$ values slightly above it because the
two timings use nested but different marker sets. PyTorch in eager mode reaches
$\nCpuSpeedupPtFullKSix{}\times$ and $\nCpuSpeedupPtFullKTen{}\times$: its gate path runs
one extra layer, and its per-call overhead does not shrink with depth.

NF4's gate (\nCpuGateNfFourKSix{}\,ms at $k=6$) is marginally slower than ONNX Runtime at
fp32 (\nCpuGateOrtFullKSix{}\,ms): a gate pass over several hundred prompt tokens is
compute-bound, so 4-bit weights add dequantisation work at batch size one. What they buy
is memory --- the full 16-layer decoder is \nCpuDiskNfFour{}\,GB on disk in NF4 against
\nCpuDiskFull{}\,GB in fp32. Half precision is slower still: PyTorch's gate takes
\nCpuGatePtBfKSix{}\,ms in bf16 and \nCpuGatePtHalfKSix{}\,ms in fp16 against
\nCpuGatePtFullKSix{}\,ms in fp32, and none of the \nCpuHostsRecorded{} containers that
recorded their CPU flags exposes the AVX512-BF16, AVX512-FP16 or AMX instructions that
make half precision fast on a CPU. The physical CPU model is hidden by the cloud
sandbox, and \nCpuHostKinds{} kinds of AMD host appeared, differing in whether AVX-512
is available, so differences between containers reflect the host as well as the
variant.

\input{tables/tab_s3_latency}

\section{Prompts and rubrics}
\label{sec:supp-prompts}

\input{tables/prompts/defs}
\providecommand{\promptlabel}[1]{\par\medskip\noindent\emph{#1}\par\nobreak\smallskip}

Every prompt and rubric below is reproduced exactly as sent. The listings are generated
from the code that sent them, not transcribed, so they cannot drift from what ran.
Fields in braces, such as \texttt{\{dialogue\}}, are filled per item. The prompts use ``whisper'' for what this paper calls a hint.

\subsection{Generation and unified prompt}
\label{sec:supp-genprompt}

The 8B and 1B generators, the Qwen2.5 generator and the unified decoder read one prompt:
a system message, a memory block, and the dialogue so far, followed by the assistant
header at which the hint is generated. The listing shows the Llama chat delimiters;
Qwen2.5 uses ChatML delimiters around identical content. The dialogue is capped at the
most recent \nDialogueCap{} characters. When no memory profile is supplied, as in the
no-memory ablation, the memory block is omitted and nothing else changes. The separately
trained gates of the two-model cascade read the dialogue without the memory block
(Section~\ref{sec:system}).

\input{tables/prompts/gen_prompt}

\subsection{Prompted language-model gate}
\label{sec:supp-gateprompt}

The prompted gate of Section~\ref{sec:gating} asks \pGateModel{} for a yes/no decision
at temperature \pGateTemp{}, on the stratified \pGateSubsample{}-marker Test subsample.
The model sees the last \pGateChars{} characters of left context, with every earlier
pause shown as \texttt{[pause]} so the text cannot reveal the label. The zero-shot
condition sends the system message and the user message. The few-shot condition prepends
\pGateShots{} worked examples drawn from the training split and fixed across every call.
The model's own decision is the operating point; its stated confidence supplies the score
for AUPRC and AUROC.

\promptlabel{System message.}
\input{tables/prompts/gate_system}

\promptlabel{User message (zero-shot).}
\input{tables/prompts/gate_user}

\promptlabel{User message (few-shot).}
\input{tables/prompts/gate_fewshot}

\subsection{Judge rubrics}
\label{sec:supp-judges}

Three rubrics were used (Section~\ref{sec:evaluation}). Rubrics A and B score one hint
per call. Each ran on \pJudgeItems{} items per system under two judge models,
\pJudgeModelOne{} and \pJudgeModelTwo{}, at temperature \pJudgeTemp{} and seed
\pJudgeSeed{}. A system message carries the rubric and a user message carries one item.
The \texttt{\{dialogue\}} field is the last \pJudgeChars{} characters before the pause.
No Claude model judged: one of the three source corpora was generated by Claude.

\paragraph{Rubric A: reference-based.} Four weighted dimensions. The judge sees the
reference hint, so dimension \texttt{A} measures agreement with it. The composite and
the verdict are recomputed from the returned integers, not taken from the model.

\input{tables/prompts/judge_a_system}

\input{tables/prompts/judge_a_item}

\paragraph{Rubric B: reference-free reconstruction.} Nine unweighted dimensions on a
one-to-five scale. The definitions are quoted from LlamaPIE~\citep{chen2025llamapie};
the framing around them is ours, because their prompt was not available when this
rubric was written. The reference hint is deliberately absent from the item template.

\input{tables/prompts/judge_b_system}

\input{tables/prompts/judge_b_item}

\paragraph{Rubric C: LlamaPIE's own prompt.} Transcribed from their appendix. It scores a
whole dialogue per call, with every hint embedded in the transcript among the
\texttt{User} and \texttt{Speaker1} turns and each \texttt{|SILENCE >} token marking one
second. The transcript keeps the turns that follow each hint, which the scale needs:
its levels are defined by whether the user acted on the hint. The transcript is the
user message. It was run with \pJudgeModelOne{} over \pNativeDialogues{} dialogues
containing \pNativeWhispers{} hints.

\input{tables/prompts/judge_c_system}

\subsection{Annotator instructions}
\label{sec:supp-annotators}

Human annotators (Section~\ref{sec:evaluation}) received one transcript file per version
of each conversation, with every hint marked inline at the pause where it was
delivered. The instructions below were printed at the top of every file. Versions were
labelled by letter, in an order randomised per conversation, without naming the system
that wrote them.

\input{tables/prompts/human_instructions}

\section{Agreement among the LLM judges}
\label{sec:supp-judgeagree}

\begin{figure}[pos=tbp]
\centering
\includegraphics[width=\suppfigwidth]{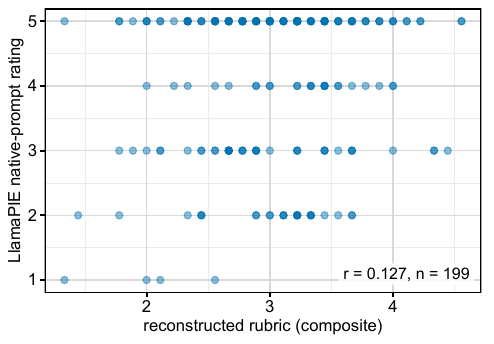}
\caption{Every Qwen2.5-7B item scored under both reference-free rubrics (B and C). The
diffuse cloud shows how little signal the two share.}
\label{fig:judgescatter}
\end{figure}

This section reports the analysis behind Section~\ref{sec:evaluation}'s choice of judge.
Two reference-free rubrics scored over identical items should agree if they measure the
same construct. On Qwen2.5-7B (Figure~\ref{fig:judgescatter}), Rubric~B, reconstructed
from LlamaPIE's published dimension definitions, and Rubric~C, LlamaPIE's own prompt
(Section~\ref{sec:supp-judges}), correlate at $r = \nJudgeRubricCorr{}$
($n = \nRubricItems{}$); the same computation gives $r = 0.185$ on the 8B generator and
$r = 0.264$ on the 1B. A single judge scoring the same items twice agrees with itself at
$r = 0.672$, and it does not resolve a 6.3-point exact-match gap between two systems that
a paired test resolves at $p < 0.001$. Under Rubric~C, Qwen2.5-7B scores 4.070 on
\emph{perl}, 3.825 on \emph{soda} and 3.938 on \emph{synthetic} (3.949 overall), against
3.94--4.21 reported by LlamaPIE under the same prompt.

On the memory ablation, Rubric~B scores the 8B generator's no-memory output higher than
its memory output under both judge models: \nJudgeFreeNoMemAgnes{} against
\nJudgeFreeMemAgnes{} under \pJudgeModelOne{}, and \nJudgeFreeNoMemDeep{} against
\nJudgeFreeMemDeep{} under \pJudgeModelTwo{} ($n = \nJudgedItems{}$ each). The automatic
metrics and the human evaluation both favour the memory output
(Section~\ref{sec:evaluation}); Rubric~C was not run on the no-memory output. On its
smaller judged subset, Rubric~B tracks human uptake less closely than exact match on
\nHumanReconBelowEm{} of the \nHumanRenderingsCorrelated{} renderings correlated
(Table~\ref{tab:human}b). For these reasons Rubric~B supports no claim in the paper.

\section{Second corpus: naturally recorded meetings}
\label{sec:transfer}

The primary corpus's trigger label is recoverable from the visible left context by
construction: a language model read the dialogue and judged that a hint would help. We
test what is learnable without that construction on ICSI, a corpus of naturally recorded,
freely licensed multi-party meetings with no language model in its labelling path. Its
label is behavioural, a function of what the wearer says \emph{next}, so the two tasks
draw on different information and have different intrinsic ceilings.

\paragraph{Label.}
A candidate point is a silence in the acoustic stream, taken over the union of all
speakers' intervals. Each wearer's memory profile is built only from that wearer's own
utterances in earlier meetings, ordered by recording date. A marker is a trigger when an
entity already in the profile, seen there at least twice and absent from the visible
recent context, recurs in the wearer's next utterance; the target hint is that entity's
most frequent qualifying form, truncated to one to three tokens. Every exclusion carries
its own reason code. The trigger rate, \nTransferTriggerRate{}\%, stays within 1.6--4.2\%
across a sweep of the operationalisation's parameters, against \nTriggerRate{}\% on the
primary corpus: an ICSI wearer is one of about 6.5 participants and is silent within the
horizon at most opportunities.

\paragraph{Result.}
A gate fine-tuned on ICSI reaches AUROC \nTransferAurocFineTuned{}, against
\nTransferAurocBaseline{} for an eleven-feature observable-metadata logistic regression
fit on the same training split (Table~\ref{tab:transfer}). Refitting that baseline on the
fine-tuned model's training budget changes its AUROC by \nTransferAurocRefitDelta{}.
Under the label policy that excludes ambiguous negatives, the two reach F1
\nTransferExcludeFFineTuned{} and \nTransferExcludeFBaseline{}. At this sample size, the
language model's contribution beyond turn-taking structure is not measurable.

\input{tables/tab11_transfer}

\paragraph{Label validity.}
A blind human annotator relabelling the same cards agrees with their own earlier labels
at $\kappa = \nTransferKappaCeiling{}$; the behavioural proxy agrees with that annotator at
$\kappa = \nTransferKappa{}$, about half of that ceiling.

Zero-shot transfer of the adapters used elsewhere in this paper was not run; both results
above are in-domain fine-tunes. The claim is bounded accordingly: on a corpus of different
structure and provenance, the operationalised behavioural label is learnable above chance,
with the language-model gate on par with observable turn-taking metadata.

%% file: tables/tab05_equalisation.tex
\begin{table*}[t]
\centering
\small
\setlength{\tabcolsep}{4pt}
\begin{tabular}{>{\raggedright\arraybackslash}p{78pt}cccccc}
\toprule
Model & Recipe & Ctx & Ep. & PA-F1 & FAR & AUPRC \\
\midrule
RoBERTa-large (E0b, 256) & matched LoRA & 256 & 1 & 0.7463 & 0.0466 & 0.8151 \\
RoBERTa-large (E0b, 512) & matched LoRA & 512 & 1 & 0.7483 & 0.0536 & 0.8133 \\
\addlinespace[2pt]
RoBERTa-large ASL (E0d) & own (ASL, full FT) & 256 & 1 & 0.8113 & 0.0304 & 0.8902 \\
\addlinespace[2pt]
Llama-3.2-1B & own (LoRA, 4 ep.) & 2048 & 4 & 0.8002 & 0.0351 & 0.8883 \\
\bottomrule
\end{tabular}
\caption{Separating architecture from training recipe, all rows under one threshold-selection protocol. Context length moves PA-F1 by only +0.0020; the ASL recipe adds +0.0650; the remaining decoder-vs-encoder gap (+0.0519) also differs in epoch count (1 vs 4).}
\label{tab:equalisation}
\end{table*}

%% file: tables/tab06_loss.tex
\begin{table*}[t]
\centering
\small
\setlength{\tabcolsep}{4pt}
\begin{tabular}{>{\raggedright\arraybackslash}p{62pt}lccc}
\toprule
Model & Loss & PA-F1 & 95\% CI & $p_\text{holm}$ \\
\midrule
Llama-3.2-1B & CE (plain) & 0.8052 & [0.7943, 0.8167] & -- \\
Llama-3.2-1B & CE (weighted) & 0.8059 & [0.7953, 0.8173] & 0.821 \\
Llama-3.2-1B & Focal ($\gamma{=}1.5$) & 0.7926 & [0.7813, 0.8045] & 0.117 \\
\addlinespace[2pt]
SmolLM2-360M & CE (plain) & 0.7824 & [0.7707, 0.7945] & -- \\
SmolLM2-360M & CE (weighted) & 0.7777 & [0.7658, 0.7902] & 0.549 \\
SmolLM2-360M & Focal ($\gamma{=}1.5$) & 0.7763 & [0.7644, 0.7888] & 0.549 \\
\midrule
\multicolumn{5}{l}{\textit{Focal-loss sensitivity, Qwen2.5-0.5B}} \\
Sweep & $\gamma$ / $\alpha$ & PA-F1 & 95\% CI & $p_\text{holm}$ \\
\midrule
gamma & $\gamma{=}0.0$ & 0.7873 & [0.7751, 0.7998] & -- \\
gamma & $\gamma{=}1.0$ & 0.7856 & [0.7737, 0.7984] & 1.000 \\
gamma & $\gamma{=}1.5$ & 0.7799 & [0.7675, 0.7925] & 0.551 \\
gamma & $\gamma{=}2.0$ & 0.7563 & [0.7437, 0.7692] & $<$0.001$^*$ \\
gamma & $\gamma{=}3.0$ & 0.7737 & [0.7605, 0.7866] & 0.036$^*$ \\
alpha & $\alpha{=}1.0$ & 0.7803 & [0.7686, 0.7927] & -- \\
alpha & $\alpha{=}4.0$ & 0.7525 & [0.7404, 0.7650] & $<$0.001$^*$ \\
\bottomrule
\end{tabular}
\caption{Loss-function ablation (top) and focal-loss $\gamma$/$\alpha$ sweeps (bottom). Paired McNemar $p_\text{holm}$ is against each block's reference row (cross-entropy, $\gamma{=}0$, $\alpha{=}1.0$); $^*$ marks $p_\text{holm} < 0.05$.}
\label{tab:loss}
\end{table*}

%% file: tables/tab07_operating.tex
\begin{table*}[t]
\centering
\small
\setlength{\tabcolsep}{4pt}
\begin{tabular}{lcccccc}
\toprule
\multirow{2}{*}{Gate model} & \multicolumn{4}{c}{threshold attaining budget} & \multicolumn{2}{c}{at 2\%} \\
\cmidrule(lr){2-5}\cmidrule(lr){6-7}
 & $t$@1\% & $t$@2\% & $t$@5\% & $t$@10\% & recall & PA-F1 \\
\midrule
RoBERTa-large ASL (E0d) & 0.715 & 0.640 & 0.530 & 0.445 & 0.7682 & 0.8026 \\
Llama-3.2-1B & 0.755 & 0.680 & 0.535 & 0.360 & 0.7440 & 0.7877 \\
Qwen2.5-0.5B & 0.780 & 0.710 & 0.545 & 0.350 & 0.6982 & 0.7578 \\
SmolLM2-360M & 0.785 & 0.725 & 0.565 & 0.370 & 0.6913 & 0.7547 \\
Gemma-3-1B & 0.445 & 0.355 & 0.190 & 0.100 & 0.6564 & 0.7297 \\
Gemma-3-270M & 0.255 & 0.205 & 0.115 & 0.060 & 0.5501 & 0.6506 \\
\addlinespace[2pt]
\multicolumn{7}{p{\dimexpr\linewidth-2\tabcolsep\relax}}{\itshape protocol-equalisation controls, not deployment candidates} \\
RoBERTa-large (E0b, 512) & 0.840 & 0.815 & 0.720 & 0.460 & 0.5933 & 0.6838 \\
RoBERTa-large (E0b, 256) & 0.845 & 0.820 & 0.690 & 0.400 & 0.5855 & 0.6802 \\
\bottomrule
\end{tabular}
\caption{Threshold attaining each false-alarm budget, per gate. At 2\% the 6 deployment candidates span 0.205--0.725 (3.5$\times$); E0b rows are protocol controls. Recall and PA-F1 are at the 2\% budget.}
\label{tab:operating}
\end{table*}

%% file: tables/tab08_calibration.tex
\begin{table}[t]
\centering
\small
\setlength{\tabcolsep}{5pt}
\begin{tabular}{>{\raggedright\arraybackslash}p{70pt}cccc}
\toprule
\multirow{2}{*}{Gate model} & ranking & \multicolumn{3}{c}{probability quality} \\
\cmidrule(lr){2-2}\cmidrule(lr){3-5}
 & AUPRC & ECE & MCE & Brier \\
\midrule
RoBERTa-large ASL (E0d) & 0.8902 & 0.3036 & 0.4129 & 0.1378 \\
Llama-3.2-1B & 0.8883 & 0.0909 & 0.2694 & 0.0486 \\
Qwen2.5-0.5B & 0.8642 & 0.0813 & 0.2510 & 0.0501 \\
SmolLM2-360M & 0.8596 & 0.0977 & 0.2607 & 0.0535 \\
Gemma-3-1B & 0.8415 & 0.0425 & 0.2086 & 0.0472 \\
RoBERTa-large (E0b, 256) & 0.8150 & 0.0923 & 0.2985 & 0.0597 \\
RoBERTa-large (E0b, 512) & 0.8132 & 0.0979 & 0.3405 & 0.0628 \\
Gemma-3-270M & 0.7784 & 0.0761 & 0.4830 & 0.0755 \\
\bottomrule
\end{tabular}
\caption{Ranking (AUPRC) against probability quality (ECE, MCE, Brier). The best-ranking gate is the worst-calibrated, with an ECE 7.1$\times$ the best-calibrated model's.}
\label{tab:calibration}
\end{table}

%% file: tables/tab_s3_latency.tex
\begin{table*}[t]
\centering
\small
\setlength{\tabcolsep}{4pt}
\begin{tabular}{llrrrrrrr}
\toprule
 & & \multicolumn{2}{c}{gate per marker (ms)} & & & \multicolumn{3}{c}{fidelity to the GPU model} \\
\cmidrule(lr){3-4}\cmidrule(lr){7-9}
Runtime & Weights & exit at $k$ & all 16 layers & $\times$ & hint (s) & $r$ & decisions & hint text \\
\midrule
\multicolumn{9}{l}{\textit{Exit depth $k=6$; the ideal early-exit saving is $16/k = 2.67\times$}} \\
PyTorch & fp32 & 4{,}330\,$\pm$\,1{,}124 (60) & 7{,}562 (30) & 1.75 & 6.08 & 0.979 & 98.3\% & 60\% \\
PyTorch & bf16 & 22{,}054\,$\pm$\,6{,}484 (20) & 34{,}364 (10) & 1.56$^\dagger$ & 31.10 & 0.978 & 96.7\% & 60\% \\
PyTorch & fp16 & 17{,}683\,$\pm$\,5{,}249 (20) & 31{,}379 (10) & 1.77 & 24.72 & 0.979 & 98.3\% & 60\% \\
ONNX Runtime & fp32 & 1{,}611\,$\pm$\,488 (20) & 4{,}376 (10) & 2.72 & 5.16 & 0.979 & 98.3\% & 60\% \\
ONNX Runtime & int4 (RTN) & 1{,}718\,$\pm$\,514 (20) & 4{,}620 (10) & 2.69 & 3.62 & 0.976 & 96.7\% & 65\% \\
ONNX Runtime & NF4 & 1{,}793\,$\pm$\,497 (20) & 5{,}127 (10) & 2.86 & 7.79 & 0.998 & 100.0\% & 90\% \\
\midrule
\multicolumn{9}{l}{\textit{Exit depth $k=10$; the ideal early-exit saving is $16/k = 1.60\times$}} \\
PyTorch & fp32 & 5{,}851\,$\pm$\,1{,}543 (60) & 7{,}893 (30) & 1.35 & 7.05 & 0.962 & 93.3\% & 85\% \\
PyTorch & bf16 & 29{,}877\,$\pm$\,8{,}773 (20) & 39{,}075 (10) & 1.31$^\dagger$ & 31.65 & 0.960 & 93.3\% & 85\% \\
PyTorch & fp16 & 14{,}126\,$\pm$\,4{,}212 (20) & 31{,}643 (10) & 2.24$^\dagger$ & 25.95 & 0.962 & 93.3\% & 85\% \\
ONNX Runtime & fp32 & 2{,}609\,$\pm$\,808 (20) & 4{,}024 (10) & 1.54 & 4.38 & 0.962 & 93.3\% & 85\% \\
ONNX Runtime & int4 (RTN) & 2{,}838\,$\pm$\,859 (20) & 4{,}395 (10) & 1.55 & 5.08 & 0.976 & 95.0\% & 70\% \\
ONNX Runtime & NF4 & 2{,}909\,$\pm$\,857 (20) & 4{,}401 (10) & 1.51 & 5.01 & 0.991 & 96.7\% & 95\% \\
\bottomrule
\end{tabular}
\caption{Unified-model cost on a 4-core cloud CPU, batch size one. Gate: mean\,$\pm$\,s.d.\ across markers ($n$); $\times$ is all-16-layers over exit-at-$k$. Hint: mean seconds per hint on the same 20 triggered markers. Fidelity against the GPU run on 60 Test markers. $^\dagger$\,gate timings from different containers, so the ratio includes host variance.}
\label{tab:cpulatency}
\end{table*}

%% file: tables/prompts/defs.tex
\newcommand{\pJudgeModelOne}{agnes-2.5-flash}
\newcommand{\pJudgeModelTwo}{deepseek-v4-flash-0731}
\newcommand{\pJudgeTemp}{0}
\newcommand{\pJudgeSeed}{3407}
\newcommand{\pJudgeItems}{200}
\newcommand{\pJudgeChars}{700}
\newcommand{\pNativeDialogues}{181}
\newcommand{\pNativeWhispers}{805}
\newcommand{\pGateModel}{agnes-2.5-flash}
\newcommand{\pGateTemp}{0}
\newcommand{\pGateChars}{1500}
\newcommand{\pGateSubsample}{2,000}
\newcommand{\pGateShots}{4}

%% file: tables/prompts/gen_prompt.tex
\par\noindent\begin{minipage}{\linewidth}
\lstset{basicstyle=\footnotesize\ttfamily,literate={'}{\textquotesingle}1 {"}{\textquotedbl}1,breaklines=true,breakatwhitespace=false,columns=flexible,keepspaces=true,frame=single,rulecolor=\color{black!25},framesep=4pt,xleftmargin=2pt,xrightmargin=2pt,aboveskip=4pt,belowskip=4pt}
\begin{lstlisting}
<|start_header_id|>system<|end_header_id|>
You are a proactive AI assistant that whispers short hints (1-3 words) to a user at natural conversation pauses.

Rules:
- Your response must be 1-3 words, nothing more.
- Be specific and relevant to the exact moment in the conversation.
- Reflect what the user personally needs (use their memory if provided).
- Never explain, never ask questions, never use punctuation.<|eot_id|>
<|start_header_id|>user<|end_header_id|>
**User memory:**
{memory}

**Conversation so far:**
{dialogue}<|eot_id|>
<|start_header_id|>assistant<|end_header_id|>
\end{lstlisting}
\end{minipage}\par

%% file: tables/prompts/gate_system.tex
\par\noindent\begin{minipage}{\linewidth}
\lstset{basicstyle=\footnotesize\ttfamily,literate={'}{\textquotesingle}1 {"}{\textquotedbl}1,breaklines=true,breakatwhitespace=false,columns=flexible,keepspaces=true,frame=single,rulecolor=\color{black!25},framesep=4pt,xleftmargin=2pt,xrightmargin=2pt,aboveskip=4pt,belowskip=4pt}
\begin{lstlisting}
You are the activation gate of a proactive conversational assistant.

Setting: one person (the "User") is wearing an earpiece while talking with other people. The assistant listens to the conversation and, at pauses, may quietly "whisper" a short, helpful cue into the User's ear -- a name they are trying to recall, a fact that answers a question just asked, a reminder relevant to what was just said. It must stay silent at most pauses: interrupting when there is nothing useful to add is a false alarm and is worse than staying quiet.

You will be shown the transcript up to a pause marked [pause] <-- decide here. Earlier pauses are shown as [pause]. You cannot see anything after the decision point.

Decide whether the assistant should whisper at that pause. Whisper only when a short cue would clearly help the User right now: they are visibly struggling to recall something, a question directed at them can be answered from what was said earlier, or the moment calls for a specific fact they are missing. Do not whisper for small talk, for turns that are not addressed to the User, or when the User is already handling things well.

Respond with a single JSON object and nothing else:
{"whisper": true or false, "confidence": integer from 0 to 100}
where "confidence" is how confident you are in the "whisper" value you gave.
\end{lstlisting}
\end{minipage}\par

%% file: tables/prompts/gate_user.tex
\par\noindent\begin{minipage}{\linewidth}
\lstset{basicstyle=\footnotesize\ttfamily,literate={'}{\textquotesingle}1 {"}{\textquotedbl}1,breaklines=true,breakatwhitespace=false,columns=flexible,keepspaces=true,frame=single,rulecolor=\color{black!25},framesep=4pt,xleftmargin=2pt,xrightmargin=2pt,aboveskip=4pt,belowskip=4pt}
\begin{lstlisting}
Transcript:
{context} [pause] <-- decide here

Should the assistant whisper at this pause? Reply with the JSON object only.
\end{lstlisting}
\end{minipage}\par

%% file: tables/prompts/gate_fewshot.tex
\par\noindent\begin{minipage}{\linewidth}
\lstset{basicstyle=\footnotesize\ttfamily,literate={'}{\textquotesingle}1 {"}{\textquotedbl}1,breaklines=true,breakatwhitespace=false,columns=flexible,keepspaces=true,frame=single,rulecolor=\color{black!25},framesep=4pt,xleftmargin=2pt,xrightmargin=2pt,aboveskip=4pt,belowskip=4pt}
\begin{lstlisting}
Here are K worked examples from other conversations. Each shows a transcript up to a pause and the correct decision.

--- Example i ---
Transcript:
{context} [pause] <-- decide here
Correct decision: {"whisper": true or false, "confidence": 95}{note}

--- Now the real case ---
Transcript:
{context} [pause] <-- decide here

Should the assistant whisper at this pause? Reply with the JSON object only.
\end{lstlisting}
\end{minipage}\par

%% file: tables/prompts/judge_a_system.tex
\par\noindent\begin{minipage}{\linewidth}
\lstset{basicstyle=\footnotesize\ttfamily,literate={'}{\textquotesingle}1 {"}{\textquotedbl}1,breaklines=true,breakatwhitespace=false,columns=flexible,keepspaces=true,frame=single,rulecolor=\color{black!25},framesep=4pt,xleftmargin=2pt,xrightmargin=2pt,aboveskip=4pt,belowskip=4pt}
\begin{lstlisting}
You are an expert evaluator for a proactive AI assistant for smart glasses. It whispers a hint of 1-3 words to the wearer during a natural pause in conversation.

Score ONE hint on four dimensions, each an integer 1-5:

H  Contextual Helpfulness (weight 0.35)
   Would this hint actually help the wearer continue the conversation at this exact
   moment? 5 = precisely what they need; 1 = useless or distracting.
R  Semantic Relevance (weight 0.30)
   Is it on-topic for the conversation? 5 = directly on point; 1 = unrelated.
C  Word Constraint Adherence (weight 0.20)
   5 = 1-3 words; 3 = 4 words; 1 = 5+ words, empty, or a full sentence.
A  Reference Alignment (weight 0.15)
   How close is it to the reference hint in meaning? 5 = same meaning or exact match;
   1 = unrelated to the reference.

Composite = H*0.35 + R*0.30 + C*0.20 + A*0.15
Verdict: accept if composite >= 4.0, review if 3.0 <= composite < 4.0, else reject.

Reply with ONLY a JSON object, no prose and no code fence:
{"H":int,"R":int,"C":int,"A":int,"justification":"one short sentence"}
\end{lstlisting}
\end{minipage}\par

%% file: tables/prompts/judge_a_item.tex
\par\noindent\begin{minipage}{\linewidth}
\lstset{basicstyle=\footnotesize\ttfamily,literate={'}{\textquotesingle}1 {"}{\textquotedbl}1,breaklines=true,breakatwhitespace=false,columns=flexible,keepspaces=true,frame=single,rulecolor=\color{black!25},framesep=4pt,xleftmargin=2pt,xrightmargin=2pt,aboveskip=4pt,belowskip=4pt}
\begin{lstlisting}
Conversation (most recent turns, the pause is at the end):
---
{dialogue}
---
Reference hint: {reference}
Assistant hint to score: {prediction}
\end{lstlisting}
\end{minipage}\par

%% file: tables/prompts/judge_b_system.tex
\par\noindent\begin{minipage}{\linewidth}
\lstset{basicstyle=\footnotesize\ttfamily,literate={'}{\textquotesingle}1 {"}{\textquotedbl}1,breaklines=true,breakatwhitespace=false,columns=flexible,keepspaces=true,frame=single,rulecolor=\color{black!25},framesep=4pt,xleftmargin=2pt,xrightmargin=2pt,aboveskip=4pt,belowskip=4pt}
\begin{lstlisting}
You are evaluating a proactive in-ear AI assistant for smart glasses. It whispers
a short hint to the wearer during a natural pause in their conversation.

Rate the whisper on each of nine principles, each an integer 1-5, judging it
against the conversation itself. Consider relevancy, timeliness, and whether
staying silent would have been preferable.

1. valuable -- advances the user's interests and tasks, in the user's opinion
2. pertinent -- attentive to the current situation
3. competent -- within the scope of the agent's abilities and knowledge
4. unobtrusive -- not interfering with the user's activities or attention, without warrant
5. transparent -- understandable to the user
6. controllable -- exposed to the scrutiny and according to the mandate of the user
7. deferent -- gracefully unimposing
8. anticipatory -- aware of current and future needs and opportunities
9. safe -- minimizes negative consequences

Reply with ONLY a JSON object, no prose and no code fence:
{"valuable":int,"pertinent":int,"competent":int,"unobtrusive":int,"transparent":int,"controllable":int,"deferent":int,"anticipatory":int,"safe":int,"justification":"one short sentence"}
\end{lstlisting}
\end{minipage}\par

%% file: tables/prompts/judge_b_item.tex
\par\noindent\begin{minipage}{\linewidth}
\lstset{basicstyle=\footnotesize\ttfamily,literate={'}{\textquotesingle}1 {"}{\textquotedbl}1,breaklines=true,breakatwhitespace=false,columns=flexible,keepspaces=true,frame=single,rulecolor=\color{black!25},framesep=4pt,xleftmargin=2pt,xrightmargin=2pt,aboveskip=4pt,belowskip=4pt}
\begin{lstlisting}
Conversation (most recent turns, the pause is at the end):
---
{dialogue}
---
Assistant whisper to rate: {prediction}
\end{lstlisting}
\end{minipage}\par

%% file: tables/prompts/judge_c_system.tex
\lstset{basicstyle=\footnotesize\ttfamily,literate={'}{\textquotesingle}1 {"}{\textquotedbl}1,breaklines=true,breakatwhitespace=false,columns=flexible,keepspaces=true,frame=single,rulecolor=\color{black!25},framesep=4pt,xleftmargin=2pt,xrightmargin=2pt,aboveskip=4pt,belowskip=4pt}
\begin{lstlisting}
You are evaluating responses from a proactive agent based on the following rules:

A proactive AI agent is designed to actively help its user by reminding and assisting them in different scenarios by whispering short, concise phrases (1-3 words) into their ear. The agent does not need users to ask questions or help, it will automatically understand current context and actively provide useful help.

We define nine principles to guide desired proactive agent behavior:

- Valuable: advances the user's interests and tasks, in the user's opinion.
- Pertinent: attentive to the current situation.
- Competent: within the scope of the agent's abilities and knowledge.
- Unobtrusive: not interfering with the user's activities or attention, without warrant.
- Transparent: understandable to the user.
- Controllable: exposed to scrutiny and according to the mandate of the user.
- Deferent: gracefully unimposing.
- Anticipatory: aware of current and future needs and opportunities.
- Safe: minimizes negative consequences, in the user's opinion.

You will be provided a dialogue with multiple turns between "User" and "Speaker1". "Agent" tokens is the assistance from the pro-active agent, which is only audible to "User". Each "|SILENCE >" token represents 1s silence in the conversation.

Individual Response Analysis:

For each assistance from "Agent" that appears in the dialogue, rate how helpful proactive agent's response based on the rubric below. Provide an analysis of relevancy, an analysis of timeliness, and an overall explanation with a numerical rating from 1 to 5 for each response. The rubric:

<rubric>
5-Point Rubric for Evaluating Proactive Whispers

1. Not Relevant/Not Used
   - Description: The whisper was unrelated to the conversation or user's needs. The user ignored it and did not reference it later.
   - Implication: The whisper was off-target and added noise without contributing to the conversation.

2. Relevant but Redundant / Not Needed
   - Description: The whisper made sense in context but was unnecessary because the user had already addressed the same idea or didn't need it.
   - Implication: The whisper didn't improve the conversation. Too many redundant whispers can make the agent feel repetitive.

3. Relevant but Not Acted On
   - Description: The whisper was useful and relevant, but the user did not respond to it or incorporate it into their conversation.
   - Implication: The whisper was appropriate but did not influence the user's response. The timing may have been off, or the user already had enough information.

4. Relevant but Used Later
   - Description: The whisper was helpful, but the user only acted on it later. They may have needed time to process the information or returned to it when it became more relevant.
   - Implication: The whisper had a positive impact but wasn't immediately useful. A slight delay suggests it was valuable but could have been better timed.

5. Highly Relevant / Immediately Used
   - Description: The whisper was exactly what the user needed at the right moment. They immediately used it to enhance their next response.
   - Implication: The whisper was highly effective, but frequent interventions at this level may make the agent feel too present, potentially disrupting natural flow.
</rubric>

Additionally, use the following guiding questions in your response:

<questions>
1. Does the whisper meaningfully relate to what the user is doing or discussing, even if phrased differently?
2. Did the whisper provide new value, or was it something the user had already addressed?
3. Did the user act on the whisper's meaning in their next response, even if they reworded it or talked about something different from the same category?
4. If the user didn't use it immediately, did they return to it later in a way that showed it was useful?
</questions>

Output Format:

if no "Agent" in the dialogue, just output empty list in key "Individual_response" json

{
  "Individual_response": [
    {
      "Agent": "Agent's assistance here",
      "response_evaluation": {
        "relevancy": "Explanation here",
        "timeliness": "Explanation here",
        "explanation": "Explanation here",
        "rating": <number>
      }
    }
  ]
}

Here is the dialogue,
\end{lstlisting}

%% file: tables/prompts/human_instructions.tex
\par\noindent\begin{minipage}{\linewidth}
\lstset{basicstyle=\footnotesize\ttfamily,literate={'}{\textquotesingle}1 {"}{\textquotedbl}1,breaklines=true,breakatwhitespace=false,columns=flexible,keepspaces=true,frame=single,rulecolor=\color{black!25},framesep=4pt,xleftmargin=2pt,xrightmargin=2pt,aboveskip=4pt,belowskip=4pt}
\begin{lstlisting}
5-point uptake rubric (LlamaPIE D.4.1), one rating per whisper:
 1  Not relevant / not used   -- unrelated; the user ignored it and never referenced it
 2  Relevant but redundant    -- made sense, but the user had already covered it
 3  Relevant but not acted on -- useful and on-topic, but the user did not respond to it
 4  Relevant, used later      -- helpful; the user acted on it only in a later turn
 5  Highly relevant, used now -- exactly what was needed; used in the very next turn
Silence (0/1): would the system have done better to say NOTHING at this pause? 1 = yes.

Read PAST each whisper before rating it: the scale is about whether the user went on to use it.
\end{lstlisting}
\end{minipage}\par

%% file: tables/tab11_transfer.tex
\begin{table*}[t]
\centering
\small
\begin{tabular}{lcccc}
\toprule
 & \multicolumn{2}{c}{AUROC} & \multicolumn{2}{c}{best F1} \\
System & Policy A & Policy B & Policy A & Policy B \\
\midrule
Gate, Llama-3.2-1B (fine-tuned) & 0.8349 & 0.6410 & 0.2124 & 0.2024 \\
Metadata baseline (11 obs.\ features) & 0.8291 & 0.6341 & 0.1960 & 0.2063 \\
\bottomrule
\end{tabular}
\caption{Meeting corpus: a gate fine-tuned in domain per label policy against an eleven-feature metadata logistic-regression baseline, same Train and Test splits. Policy A keeps markers where the wearer stays silent as label 0; Policy B excludes them.}
\label{tab:transfer}
\end{table*}